\documentclass[10pt,twocolumn,letterpaper]{article}

\usepackage[pagenumbers]{cvpr}   % To force page numbers, e.g. for an arXiv version

\usepackage[utf8]{inputenc} % allow utf-8 input
\usepackage[T1]{fontenc}    % use 8-bit T1 fonts
\usepackage{url}            % simple URL typesetting
\usepackage{booktabs}       % professional-quality tables
\usepackage{amsfonts}       % blackboard math symbols
\usepackage{nicefrac}       % compact symbols for 1/2, etc.
\usepackage{microtype}      % microtypography
\usepackage{multirow}  % Required for multirow cells
\usepackage{caption}
\usepackage{float}
\usepackage{overpic}
\usepackage{bbm}
\usepackage{tabularx} % 用于X列类型
\usepackage{bbding}
\usepackage{makecell}
\usepackage{enumitem}
\usepackage{wrapfig}
\usepackage{bm}
\usepackage{pifont}
\usepackage{amsmath} 
\usepackage{listings}
\usepackage[pagebackref,breaklinks,colorlinks,allcolors=cvprblue]{hyperref}
\usepackage{stfloats}

\usepackage{placeins} % 引入 placeins 包以使用 \FloatBarrier
\usepackage{framed}
\usepackage{lipsum,mwe,cuted}
\usepackage[accsupp]{axessibility}

\definecolor{cvprblue}{rgb}{0.21,0.49,0.74}
\definecolor{lightpurple}{RGB}{200, 170, 180}
\definecolor{lightblue}{RGB}{140, 180, 170}

\title{Driving by the Rules: A Benchmark for Integrating Traffic Sign Regulations into Vectorized HD Map}
\author{
Xinyuan Chang $^1$\footnotemark[1] \and Maixuan Xue $^2$\footnotemark[1] \footnotemark[2] \and Xinran Liu $^1$
\and Zheng Pan $^1$ \and Xing Wei $^2$\footnotemark[3] 
\and 
$^1$Amap, Alibaba Group \and $^2$Xi'an Jiaotong University 
\and 
{\tt\small \{changxinyuan.cxy, tom.lxr, panzheng.pan\}@alibaba-inc.com} \\
{\tt\small xmx0809@stu.xjtu.edu.cn}, {\tt\small weixing@mail.xjtu.edu.cn}
}

\begin{document}
\maketitle
\begin{strip}
    \centering
    \includegraphics[width=0.9\textwidth]{figures/overview_new.pdf}
    \captionof{figure}{\textbf{MapDR Overview and Motivation}. The image on the left depicts the comprehensive results of HD map construction for an intersection scene, whereas the image on the right illustrates the outcomes after partitioning it into three layers. Existing online mapping methods primarily emphasize the construction of the geometric and connectivity layers, neglecting the traffic regulation layer. However, precise comprehension of traffic signs and their correlation with lanes is vital for ensuring the safety of autonomous driving.}
    \label{overview}
\end{strip}

\renewcommand\thefootnote{\fnsymbol{footnote}}
\footnotetext[1]{Equal contribution.}
\footnotetext[2]{Work done during the internship at Amap, Alibaba Group.}
\footnotetext[3]{Corresponding author.}

\begin{abstract}

Ensuring adherence to traffic sign regulations is essential for both human and autonomous vehicle navigation. While current online mapping solutions often prioritize the construction of the geometric and connectivity layers of HD maps, overlooking the construction of the traffic regulation layer within HD maps. Addressing this gap, we introduce \textbf{MapDR}, a novel dataset designed for the extraction of \textbf{D}riving \textbf{R}ules from traffic signs and their association with vectorized, locally perceived HD \textbf{Maps}. 
MapDR features over $10,000$ annotated video clips that capture the intricate correlation between traffic sign regulations and lanes.
Built upon this benchmark and the newly defined task of integrating traffic regulations into online HD maps, we provide modular and end-to-end solutions: \textbf{VLE-MEE} and \textbf{RuleVLM}, offering a strong baseline for advancing autonomous driving technology. 
It fills a critical gap in the integration of traffic sign rules, contributing to the development of reliable autonomous driving systems. Code is available at \href{https://github.com/MIV-XJTU/MapDR}{https://github.com/MIV-XJTU/MapDR.}

\end{abstract}    
\section{Introduction}
\label{sec:intro}
The rise of autonomous vehicles and intelligent transportation systems has emphasized the importance of accurate, reliable navigational data. High-Definition (HD) maps, with their detailed representations of road elements, have become essential to supporting these systems. An HD map typically consists of three primary layers: the geometric layer, connectivity layer, and traffic regulation layer~\cite{hdmapsurvey, hdmapsurvey2}, as shown in \cref{overview}. The geometric layer provides vector information, such as dividers and lane centerlines; the connectivity layer defines lane relationships to assist in path planning; and the traffic regulation layer encodes rule information (e.g., HOV lanes, bus lanes, speed-limit zones) associated with lanes, ensuring compliant driving.

A limitation of conventional HD maps is their inability to support rapid, real-time updates, making the construction of online HD maps an emerging trend. Existing methods, such as MapTR~\cite{liao2023maptr} and TopoMLP~\cite{topomlp}, address only the geometric and connectivity layers, overlooking the traffic regulation layer. As a result, autonomous driving systems currently still rely on offline maps to obtain traffic regulations, an approach that runs counter to the trend of online HD map construction.

Traffic signs serve as a "visual language" on the road, playing a critical role in defining traffic regulations. Inspired by human driving processes, the creation of traffic regulations can be divided into two stages: 1) understanding the rules conveyed by traffic signs, and 2) determining the lanes to which these rules apply. Although OpenLane-V2~\cite{wang2023openlanev2} has made strides in associating signs with lanes in online mapping, it has two primary limitations. First, it considers only directional signs, neglecting other types of regulations. Second, its label annotations are limited to sign categories, falling short of the detailed rule descriptions required for autonomous driving. A more structured description, in alignment with HD map standards, is essential for supporting comprehensive autonomous driving applications~\cite{hdmapsurvey, hdmapsurvey2}.

To bridge this gap, we introduce \textbf{MapDR}, the first dataset specifically designed for the challenging task of \textbf{integrating traffic regulations into existing online HD maps.}
MapDR focuses on complex traffic scenes with diverse signage, collecting over $10,000$ video clips across three representative cities in China, with each clip containing at least one traffic sign. In addition to vectorized representations of dividers, boundaries, and centerlines, MapDR provides structural annotations of traffic regulations and their associations with lanes, offering a comprehensive foundation for advancing online HD maps. More details of MapDR can be found in \cref{4dataset}.

Based on MapDR, in addition to the end-to-end task definition, we also divide this task into two innovative sub-tasks aimed at bolstering research in this domain: 1) rule extraction from traffic signs, and 2) rule-lane correspondence reasoning.
We introduce Vision-Language Encoder \textbf{(VLE)} and Map Element Encoder \textbf{(MEE)} to handle the interaction of multimodal data, encompassing images, texts, and vectors. Through the concatenation of these two models, integrated traffic regulations can be obtained.
Additionally, inspired by recent advancements in Multimodal Large Language Models (MLLMs), we also explore an end-to-end MLLM solution called \textbf{RuleVLM} to solve this task holistically. These approaches establish robust baselines and aim to inspire future research in this area.
For detailed descriptions of the proposed tasks and metrics, please refer to \cref{3task}. For approaches, please refer to \cref{modual_approach,end2end_approach}.

To sum up, our contributions are as follows:

\begin{itemize}[leftmargin=1cm]
\item For the first time, We introduce the novel task of extracting lane-level rules from traffic signs and integrating them into vectorized HD maps, alongside the MapDR dataset and evaluation metrics for benchmarking.
\item MapDR comprises over $10,000$ video clips and $400,000$ front-view images captured across multiple cities in China, covering diverse traffic conditions and including more than $18,000$ annotated lane-level rules. All data are newly collected and meticulously validated.
\item We propose two modeling approaches, modular and end-to-end, to address the integration of lane-level rules into vectorized HD maps, providing effective baselines for future research. \end{itemize}

\begin{figure*}[t]
  \centering
  \includegraphics[width = 0.9\textwidth]{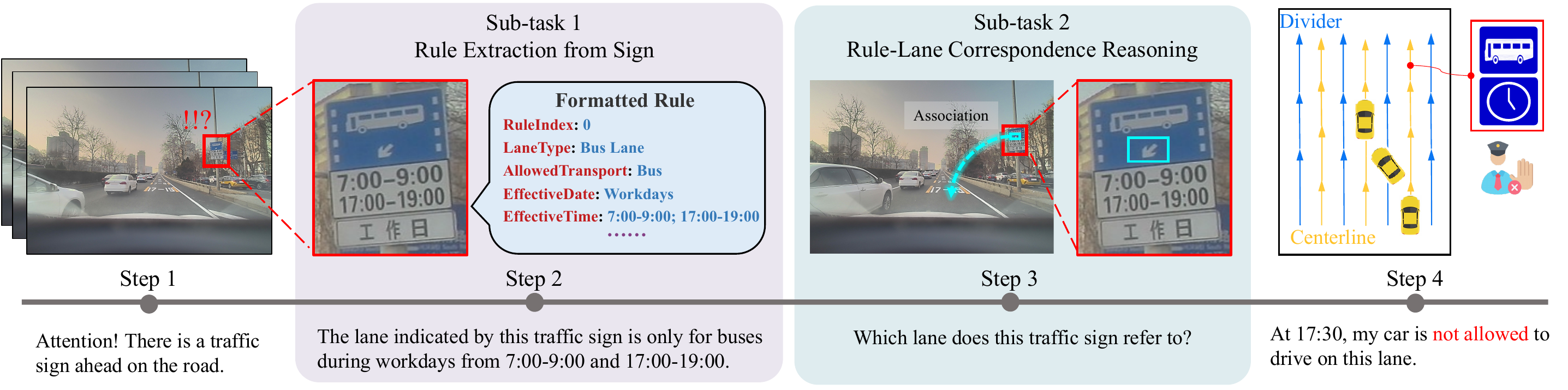}
  \caption{\textbf{Overview of the sub-tasks.} $Step \ 1$ \textasciitilde{} $Step \ 4$ shows a case of driving by the rules.  $Step \ 2$ and $Step \ 3$ demonstrates 
  the specific role of two sub-tasks, respectively.}
  \label{task defination}
\end{figure*}

\section{Related Work}
\label{sec:related_work}

\begin{table}[t]
  \centering
    \caption{\textbf{Comparison of the existing datasets.} 
    "Sign" and "Lane" denote whether the dataset focus on traffic signs and lanes. Only those annotated with formatted ("Fmt.") rules and the correspondence ("Corr.")
    between rules and lanes can form driving rules. 
    "Clip" represents whether the data is organized in the form of video clips. "$*$" denotes that these samples are not newly collected and are built upon the previous dataset.} 
  \resizebox{\columnwidth}{!}{
  \begin{tabular}{lcccccccc}
  \toprule
  \multirow{2}{*}{\textbf{Dataset}} & \multirow{2}{*}{\textbf{Sign}} & \multirow{2}{*}{\textbf{Lane}}
  & \multicolumn{2}{c}{\textbf{Driving Rules}} & \multicolumn{3}{c}{\textbf{Number of Samples}}  & \multirow{2}{*}{\textbf{Year}} \\ \cmidrule(r){4-5} \cmidrule(r){6-8}
   & & & \textbf{Fmt.} & \textbf{Corr.}  & \textbf{Image} & \textbf{Clip} & \textbf{Region} & \\
  \midrule
   nuScenes~\cite{nuscenes2019} & & \ding{51} & &  & $1400K$ & $1K$ & Worldwide & $2019$ \\
   Argoverse2~\cite{Argoverse2} & & \ding{51} & &  & $2100K$ & $1K$ & USA & $2021$ \\
   CTSU~\cite{learningtounderstandtrafficsigns} & \ding{51} & &  &  & $5K$ & /  & China & $2021$ \\
   OpenLane~\cite{chen2022persformer} & & \ding{51} & & & $200K^*$ & $1K^*$ & Worldwide & $2022$ \\
   RS10K~\cite{vtkgg} & \ding{51} & & & \ding{51}  & $10K$ & / & China & $2023$ \\
   OpenLane-V2~\cite{wang2023openlanev2} & \ding{51} & \ding{51} & & \ding{51} & $466K^*$ & $2K^*$ & Worldwide & $2023$ \\
   Waymo~\cite{waymo} & \ding{51} & \ding{51} & &  & $390K$ & $2K$ & USA & $2024$ \\
   \rowcolor{gray!15} \textbf{MapDR(ours)} & \ding{51} & \ding{51} & \ding{51} & \ding{51} & \bm{$400K$} & \bm{$10K$} & \textbf{China} & \bm{$2024$} \\
  \bottomrule
  \end{tabular}
  }
  \label{comparison of exisiting works}
\end{table}

\subsection{HD Map Construction}
HD maps construction have seen significant advancements, with a focus on traffic element perception, including lane detection and traffic sign recognition~\cite{Argoverse2, ApolloScape, nuscenes2019, RoadDetectionthroughCRF, bstld, gtsrb, bdd100k, DTLD, tt100k}. The shift towards BEV perception and vectorization for end-to-end HD maps construction has gained traction~\cite{Argoverse2, nuscenes2019, chen2022persformer}. Notable works include HDMapNet, which aggregates semantic segmentation results~\cite{li2021hdmapnet}, 
LSS~\cite{philion2020lift} estimates depth to transfer image features to BEV features, while VectorMapNet~\cite{liu2022vectormapnet} is the first end-to-end framework for sequential vector point prediction to generate HD maps without post-processing. MapTR~\cite{liao2023maptr} and its enhanced version, MapTRv2~\cite{maptrv2}, introduced a unified permutation-equivalent modeling approach and extended it to a general framework supporting centerline learning and 3D map construction. However, these efforts have largely overlooked the integration of traffic sign rules into HD maps.

\subsection{Traffic Element Association}\label{2.2}
Traffic element association aims to link elements like traffic signs with lanes. 
As demonstrated in \cref{comparison of exisiting works}, 
CTSU has initiated internal elements association to describe traffic sign in $\{key:value\}$ form, however lacking both generalization of driving rules from description and lane association~\cite{learningtounderstandtrafficsigns}.
VTKGG~\cite{vtkgg} propose to utilize a graph model for connectivity but also lacks structured expression of driving rules for motion planning and requires complex integration into HD maps, which is typically expressed in the BEV space. 
OpenLane-V2~\cite{wang2023openlanev2} advances traffic sign and lane association in BEV space, but is constrained by single-label classification for traffic sign rather than structured descriptions for fine-grained driving rules. This makes it insufficient for signs with multiple rules, which are complex but common in real scenarios.
Recent MLLM-based benchmarks~\cite{marcu2024lingoqa, qian2024nuscenesqa, sachdeva2023rank2tell, sima2023drivelm, tencent2023maplm} for autonomous driving, such as MAPLM~\cite{tencent2023maplm}, prioritize end-to-end motion planning over precise rule extraction from traffic sign, lacking evaluation for rule reasoning. 
MapDR addresses this gap by focusing on traffic sign rule extraction and lane association.

\subsection{Vision-Language Models}
Vision-Language Models (VLMs) facilitates multimodal applications by learning joint representations of vision and language data. Visual Question Answer (VQA) tasks provide answers to image-related questions~\cite{vqa}, while Visual Information Extraction (VIE) tasks extract structured information from visual and textual data~\cite{vqa, Xu2020LayoutLMPO,Xu2020LayoutLMv2MP,huang2022layoutlmv3}. In Autonomous Driving (AD), VLMs are increasingly used for comprehensive traffic scene understanding and decision-making. The field has seen various approaches, including using transformers~\cite{attention_is_all_you_need} for joint encoding~\cite{vilt, huang2022layoutlmv3}, excelling at multimodal information interaction, and independent encoders for different modalities~\cite{clip, align} that are proficient in multimodal retrieval. 
Cross-modal representation methods~\cite{ALBEF, Yu2022CoCaCC} combine these advantages, and the latest LLM-based research~\cite{li2023blip2, liu2023llava, liu2023improvedllava, liu2024llavanext} has achieved state-of-the-art results in various multimodal tasks.
Nowadays, an increasing number of methods are leveraging LLMs to achieve impressive results~\cite{gpt-driver,Drivegpt4,drivingwithllms,sima2024drivelmdrivinggraphvisual}, with works like DriveLLM~\cite{drivellm} showing significant potential in AD. However, addressing hallucination~\cite{bai2024hallucination} remains the most crucial aspect for LLM-based approaches.

\section{Task Definition : Integrating Traffic Sign Regulations into HD Maps} \label{3task}
The ability to discern rules from traffic signs and to associate them with specific lanes is pivotal for autonomous navigation. Our proposed task focuses on extracting lane-level driving rules denoted as $R = \{r_i\}_{i=1}^{m} $ and reasoning the correspondence between these rules and the centerlines $L = \{l_i \}_{i=1}^{k}$ in the local vectorized HD map, where $m$ is the number of rules and $k$ is the number of centerlines. Each rule $r_i$ is a set of pre-defined properties in $\{key:value\}$ pairs, as shown in \cref{demo}. 
The final correspondence forms a bipartite graph $G = (R \cup L,E)$, where $E\subseteq\{0,1\}^{m\times k}$ and the element $E_{ij}$ in the $i$-th row and $j$-th column of matrix $E$ represents the corresponding status between $r_i$ and $l_j$. 
In this case, we assume the entire approach as $\mathcal{F}$ and the overall process can be formulated as $G = \mathcal{F}(X, L)$, where $X = \{x_i \}_{i=1}^{n}$ represents a series of image sequences and $n$ is the number of frames.
To independently investigate the capabilities of rule extraction and correspondence reasoning within the overall task, we further divided the overall task into two sub-tasks, defined as follows.

\subsection{Rule Extraction from Traffic Sign}
As shown in $Step \ 2$ of \cref{task defination}, this sub-task involves extracting multiple rules $R = \{r_i\}_{i=1}^{m} $ from a series of image sequences $X = \{x_i \}_{i=1}^{n}$. The rule extraction model, denoted as $\cal M$, can be expressed as $R = \mathcal{M}(X)$. 

\subsection{Rule-Lane Correspondence Reasoning} \label{3.2}
As shown in $Step \ 3$ of \cref{task defination}, the reasoning process establishes the correspondence between centerlines $L = \{l_i \}_{i=1}^{k}$ and all rules $R$. We denote the correspondence reasoning model as $\mathcal{T}$, and this process can be described as $E =\mathcal{T}(R, L)$. The final reasoning result forms a bipartite graph $G = (R \cup L,E)$, which means corresponding relationships only exist between rules and centerlines.

\section{The MapDR Dataset \& Benchmark} \label{4dataset}

\begin{figure}[t]
  \centering
  \includegraphics[width = 0.9\columnwidth]{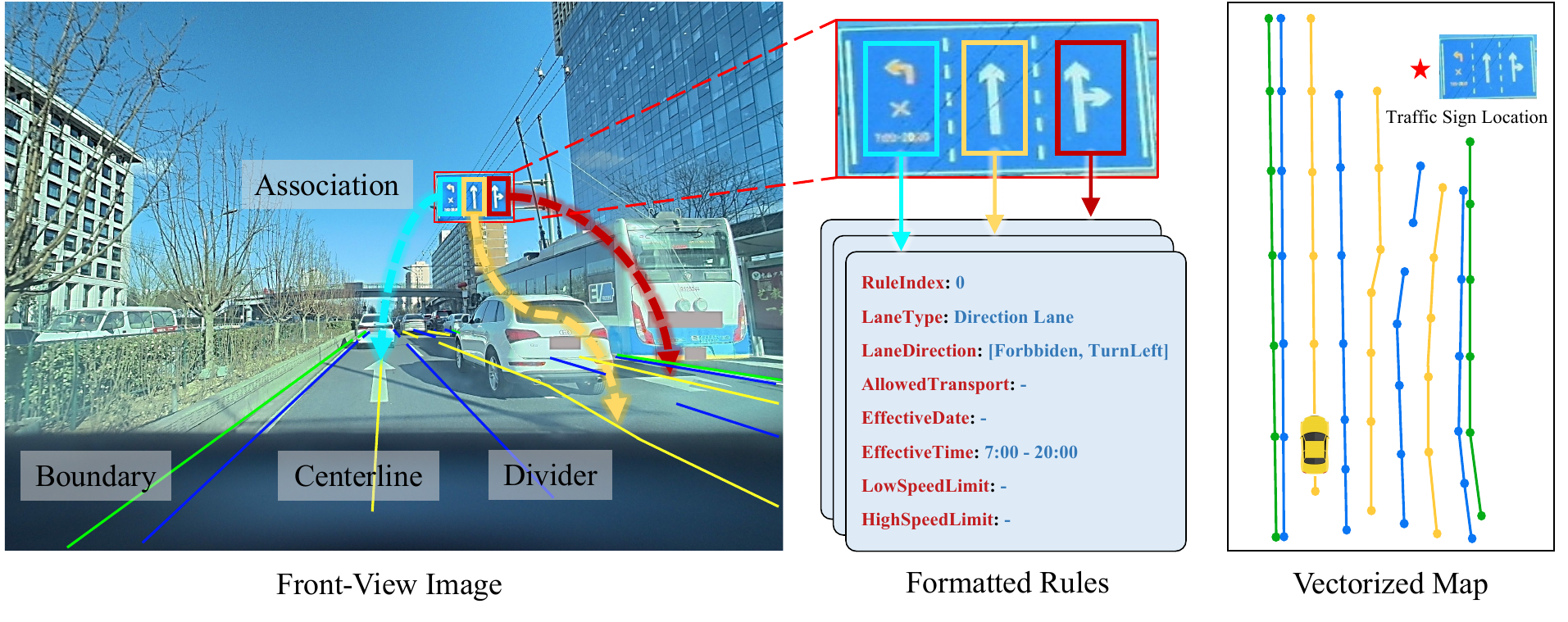}
  \caption{\textbf{Visualization of dataset demo.} Multiple lane-level rules of a single traffic sign are annotated in $\{key:value\}$ format. Directed lines indicate the correspondence between rules and particular centerlines. }
  \label{demo}
\end{figure}

We introduce the MapDR dataset, meticulously annotated with traffic sign regulations and their correspondences to lanes, as shown in \cref{demo}. The dataset encompasses a diverse range of scenarios, weather conditions, and traffic situations, with over $10,000$ traffic scene segments, $18,000$ driving rules, and $400,000$ images.
Traffic signs typically have varying textual descriptions, text layouts, and positions on the road, which add complexity to the task.

\begin{figure}[b]
  \centering
  \includegraphics[width=0.9\columnwidth]{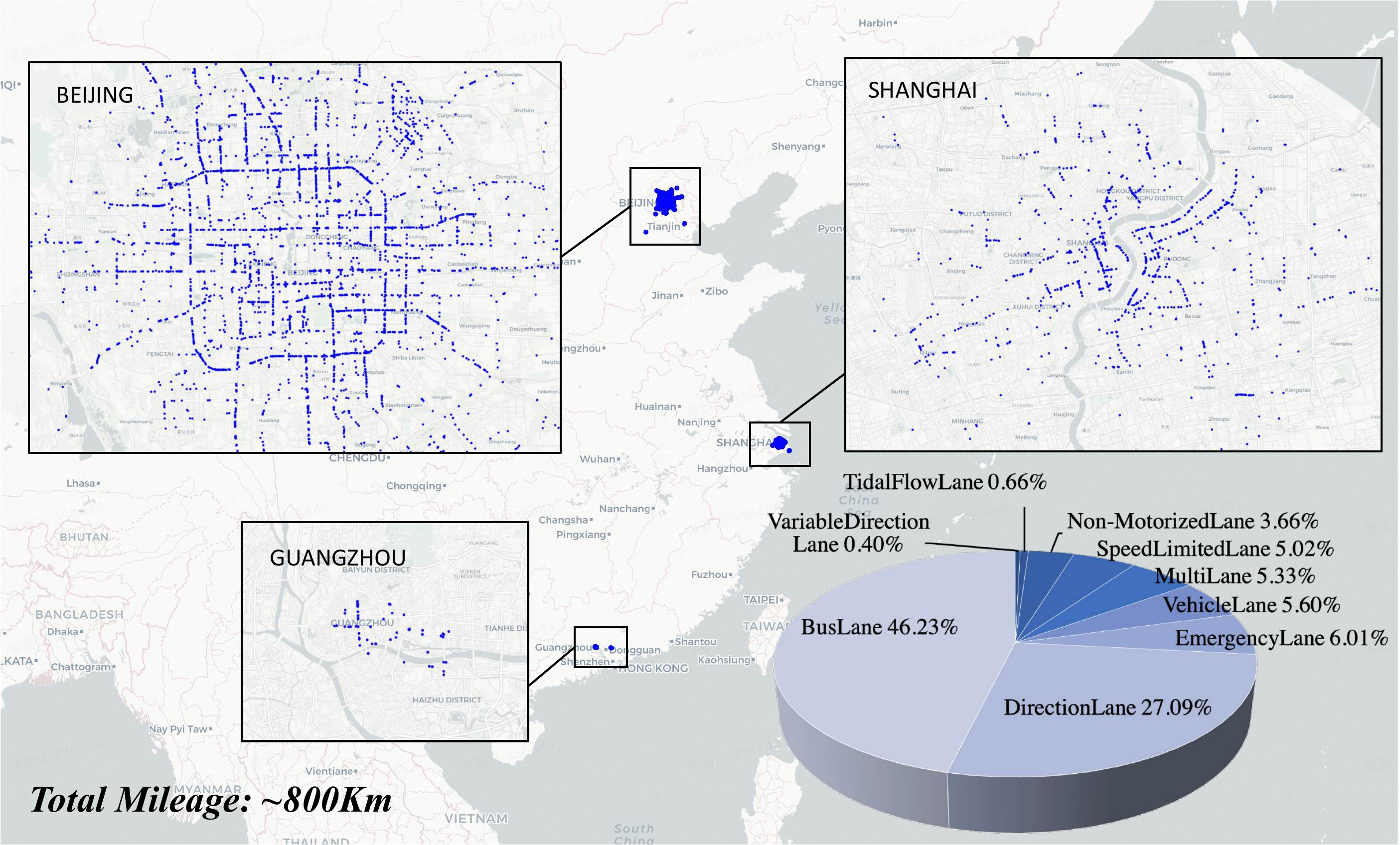}
  \caption{\textbf{Geographic location distribution of the collected traffic signs and proportions of various lane types represented in all signs.} The geographic distribution is visualized based on OpenStreetMap~\cite{osm}.}
  \label{location_map}
\end{figure}

The majority of the data originates from Beijing and Shanghai, with additional scenes from Guangzhou. \cref{location_map} illustrates the geographic spread and variety of traffic signs. The dataset reflects a natural long-tail distribution, with a prevalence of bus and direction lanes and a scarcity of tidalflow lanes.
We primarily focus on traffic signs that indicate lane-level rules, collected from cities with the most complex and diverse traffic scenarios in China, ensuring realistic and representative data.
All images have undergone privacy and safety processing to obscure license plates and faces. 
More comprehensive statistic of dataset and case demonstrations can be found in the supplement.

\subsection{Raw Data \& Annotation}
\paragraph{Raw Data.} MapDR is collected from real-world traffic scenes, each scene segment (video clip) captures front-view images within a $100m \times 100m$ area centered on the traffic sign, with a consistent resolution of $1920 \times 1240$. Each clip contains $30$ to $60$ frames, captured at $1$ frame every $2$ meters, ensuring consistent spatial intervals. Each video clip focuses on a single traffic sign and provides its position in 3D space. Camera intrinsics and poses are provided for each frame, and coordinates for each clip are transformed to distinct ENU systems. 

All vectors of local map in the target area are provided as 3D point lists, generated using our algorithm similar to MapTRv2~\cite{maptrv2}. Each lane vector has a type, such as divider, centerline, crosswalk, or boundary. For example, the centerline is defined as $L=\{l_i\}_{i=1}^{k}$, where each vector $l_i$ is composed of multiple 3D points $l_i = [p_1, \dots, p_n]$, and $p_j = (x_j, y_j, z_j)$ represents the coordinates of the current point. The pipeline of dataset production is illustrated in \cref{data_engine}, and detailed data acquisition and annotation procedures can be found in the supplement material.

\begin{figure}[t]
  \centering
  \includegraphics[width=0.9\columnwidth]{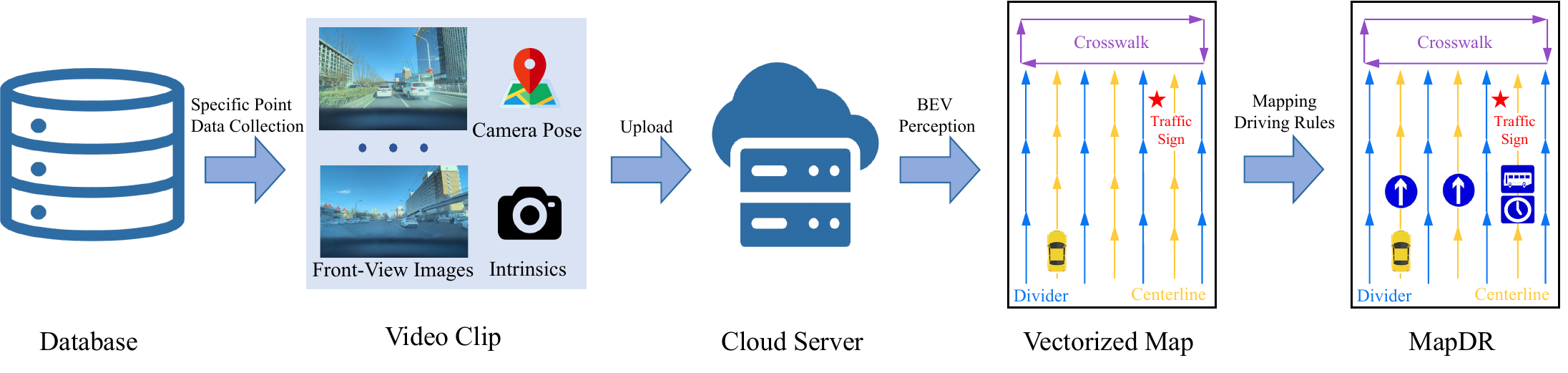}
  \caption{\textbf{Pipeline of dataset production.} The location of traffic signs are sampled from existing database then front-view images of each sign are newly collected. Vectorized map is processed in cloud server. Finally formatted rules and correspondence between rules and centerlines are annotated and organized as MapDR.}
  \label{data_engine}
\end{figure}

\paragraph{Formatted Rules.} Each video clip may contain multiple lane-level rules, denoted as $R$. Each rule is expressed by symbols and text on the sign, requiring interpretation. 
We reference existing HD map design and data specifications~\cite{hdmapsurvey, hdmapsurvey2,nds} to define lane-level driving rules in the JSON format.
As shown in \cref{demo}, each rule $r_i$ comprises $8$ predefined properties in the form of $\{key:value\}$ pairs.
This definition schema encompasses most scenarios and supports integration into existing autonomous driving systems.
We enclose the symbols and texts denoting each distinct rule on traffic signs with polygons and project them into 3D space as $P_i = [p_1, \dots, p_n]$, where $n$ varies. Researchers can optionally use this information to facilitate rule extraction. \label{4.1}

\paragraph{Correspondence between Rules \& Lanes.} Based on formatted rules $R$ and centerlines $L$, corresponding centerlines of each rule are annotated as shown in \cref{demo}. Therefore correspondence between rules and centerlines can be formed as a bipartite graph $G = (R \cup L,E)$, 
where $E\subseteq\{0,1\}^{|R|\times|L|}$ and the positive edges only exist between $R$ and $L$ as demonstrated in \cref{3.2}. Specifically, $E_{ij} = 1$ represents that vehicle driving on the lane with centerline $l_j$ should follow the driving rule $r_i$. 

\subsection{Evaluation Metrics}
Based on the task defined in \cref{3task}, we propose evaluation metrics for entire task as well as for two sub-tasks separately. The $F_1$ score is adopted as the benchmark for methods rank. 

\paragraph{Rule Extraction (R.E.).} Given the ground truth $R$ and predicted rules $\hat{R}$, we propose to calculate the $Precision$ ($P_{R.E.}$) and $Recall$ ($R_{R.E.}$) to evaluate the capability of rules extraction as defined in \cref{metric_re}, where $ \hat{r_i} = r_i $ represents all the properties are predicted correctly. 
\begin{equation}
  P_{R.E.} = \frac{| \hat{R} \cap R |}{|\hat{R}|} \qquad  
  R_{R.E.} = \frac{|\hat{R} \cap R | }{| R | }
  \label{metric_re}
\end{equation}

\paragraph{Correspondence Reasoning (C.R.).} Given the ground truth of correspondence bipartite graph $G = (R \cup L,E)$ and predicted graph $\hat{G} = (R \cup L,\hat{E})$, we propose to calculate $Precision$ ($P_{C.R.}$) and $Recall$ ($R_{C.R.}$) of edge set $E$ to evaluate the capability of correspondence reasoning individually. Metrics are defined as \cref{metric_cr}.
\begin{equation}
  P_{C.R.} = \frac{| \hat{E} \cap E |}{|\hat{E}|} \qquad  
  R_{C.R.} = \frac{| \hat{E} \cap E | }{| E |}
  \label{metric_cr}
\end{equation} 

\paragraph{Overall.}  
As defined in \cref{3task}, the final predicted results can be obtained through an end-to-end prediction or a combination of two sequential sub-tasks. 
Given the predicted rules, correspondence should be reasoned between $\hat{R}$ and $L$ which means the prediction of entire task is $\hat{G} = (\hat{R} \cup L, \hat{E})$ and the ground truth is consistent $G = (R \cup L, E)$. 
We evaluate $Precision$ ($P_{all}$) and $Recall$ ($R_{all}$) using the sub-graph $G^{s}$, where $G^{s} = \{ g^{s}_{ij}\}_{i=1, j=1}^{m, k}, { g^{s}_{ij} = (\{r_i, l_j\}, e_{ij})} $. In set of sub-graph $G^{s}$, $m$ is the number of rules, and $k$ is the number of centerlines. 
Furthermore, we propose the $F_1$ for the final benchmark ranking. Metrics are defined in \cref{metric_overall} and \cref{metric_f1}, The $F_1$ is derived from the calculations of $P_{all}$ and $R_{all}$.
We provide an example of calculating the \emph{Overall} metrics in the supplement material.

\begin{equation}
  P_{all} = \frac{| \hat{G^{s}} \cap G^{s} |}{|\hat{G^{s}}|} \qquad  
  R_{all} = \frac{|\hat{G^{s}} \cap G^{s}|}{|G^{s}| } \qquad 
\label{metric_overall}
\end{equation}

\begin{equation}
    F_1 = 2 \times \frac{P_{all} \times R_{all}}{P_{all} + R_{all}}
\label{metric_f1}
\end{equation}

\section{Modular Approach}\label{modual_approach}
In order to conduct a comprehensive analysis of the entire task, a modular approach is implemented in this section to address the two sub-tasks. It is demonstrated that these interconnected modules can effectively achieve the ultimate objective of integrating driving rules into the HD map.
To address the multimodal information interaction encompassing images, texts, and vectors, a \textbf{Vision-Language Encoder (VLE)} and a \textbf{Map Element Encoder (MEE)} are developed. The following sections expound upon their structures, applications, and the experimental results on MapDR.

\begin{figure*}[t]
  \centering
  \includegraphics[width=0.9\textwidth]{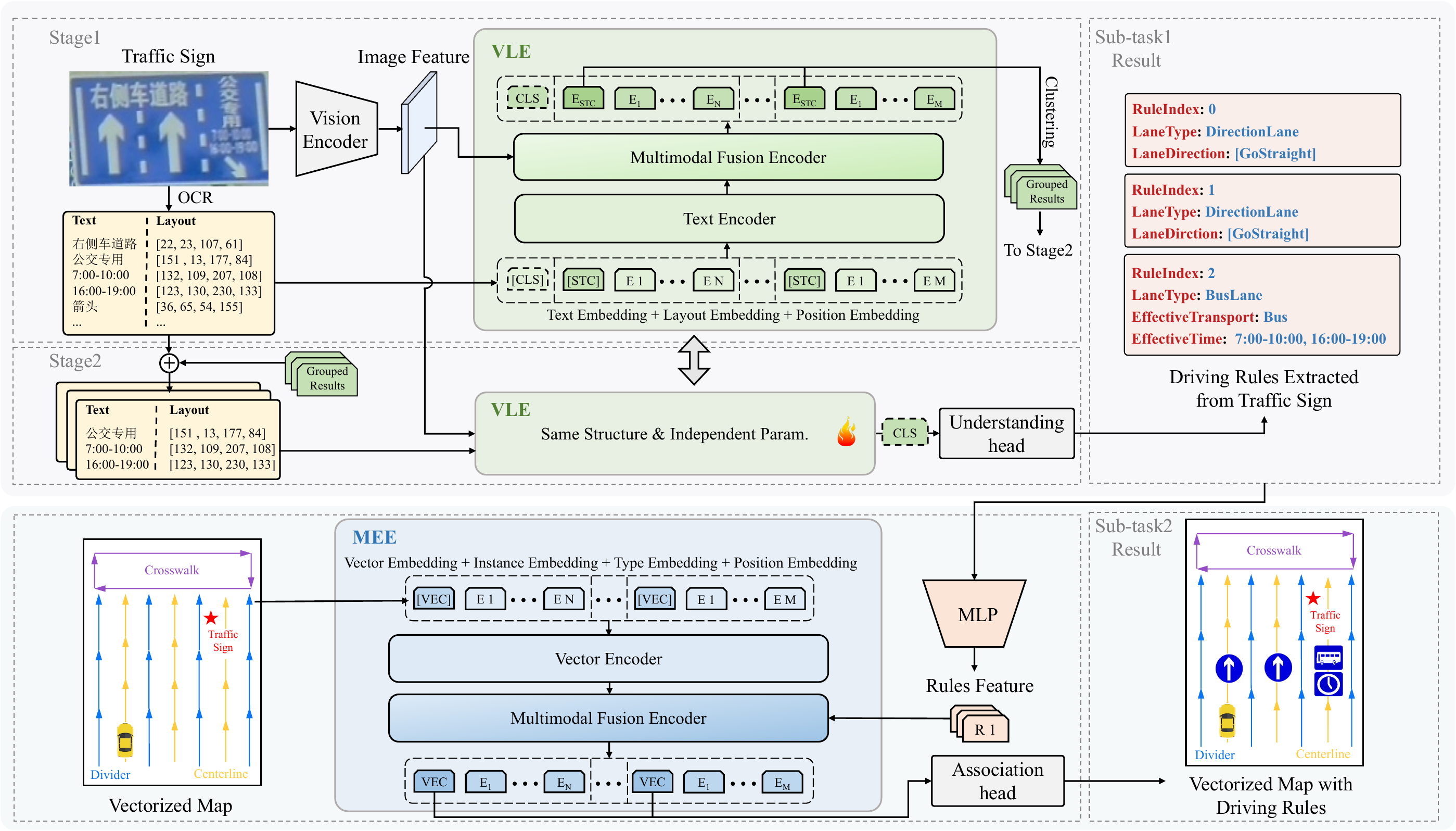}
  \caption{\textbf{Overview of the modular approach.} Entire approach can be divided into two main parts:  \textbf{Rule Extraction from Traffic Sign} (\textcolor{lightpurple}{top}) and \textbf{Rule-Lane Correspondence Reasoning} (\textcolor{lightblue}{bottom}). Rule Extraction model consists of two sequential stages with the same structure VLE but unshared parameters, and the training procedure is independent. }
  \label{overall}
\end{figure*}

\subsection{Architecture}\label{archi-all}
\paragraph{Vision-Language Encoder.}
Inspired by vision-language frameworks~\cite{ALBEF, li2022blip, clip, vilt, vlmo}, we designed a vision-language fusion model named VLE, following~\cite{ALBEF}. As shown in \cref{overall}, VLE uses ViT-b16~\cite{dosovitskiy2020vit} as the vision encoder, with the text encoder and multimodal fusion encoder each consisting of $L$ transformer layers~\cite{attention_is_all_you_need}. Each layer of the fusion encoder includes a cross-attention module for fusion~\cite{ALBEF}.
In practice, distinct rules are represented by varying numbers of symbols and texts, as shown in the OCR results in \cref{overall}. 
To address the challenge of representing variable-length input as fixed-length features, we introduce a \texttt{[CLS]} token for an entire rule and several \texttt{[STC]} tokens for sentence-level representation. The specific usage of these tokens is detailed in \cref{method-overview}.
Furthermore, we incorporate inter-instance and intra-instance attention mechanisms~\cite{maptrv2} to enhance model performance by capturing interactions and independence between and within sentences.
In addition to content, layout captures the relative positions of symbols and texts, offering important semantic meaning. To leverage this, we encode the layout using the method from~\cite{samPE} and the relative positions of characters as position embedding following~\cite{devlin2018bert}. As shown in VLE in \cref{overall}, text embedding, layout embedding, and position embedding together form the input of the text encoder.

\begin{figure}[b]
  \centering
  \includegraphics[width=0.9\columnwidth]{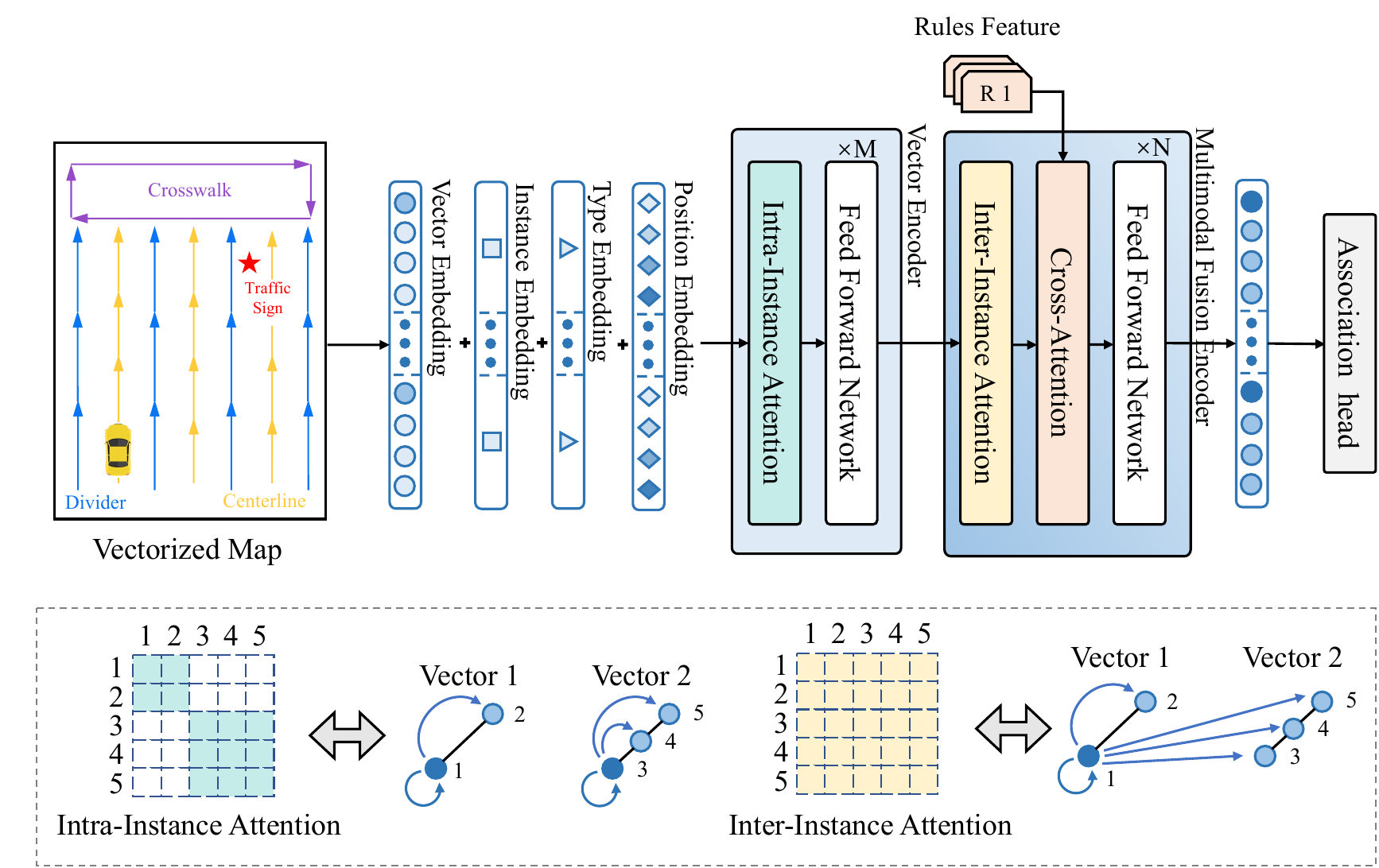}
  \caption{\textbf{Structure of MEE.} MEE serves as correspondence reasoning model. Learnable embeddings are introduced within input to enhance the representing capacity of vector types. inter \& intra-instance attention mechanisms facilitate to capture the relationships and independence of individual vectors. }
  \label{mee}
\end{figure}

\paragraph{Map Element Encoder.}\label{archi-mee}
Vectors can be represented as sequences of points, similar to words in sentences. Inspired by this, we designed MEE akin to language models~\cite{devlin2018bert}.
The MEE employs $M$ transformer layers for vector encoding and $N$ cross-attention layers for multimodal fusion.
Utilizing the method from~\cite{samPE}, points of each vector are embedded as point embedding.
To achieve a fixed-length representation, we add \texttt{[VEC]} tokens as the first token of each vector, similar to \texttt{[STC]} tokens in the VLE. 
We also introduce learnable type embedding for vector types, learnable instance embedding to distinguish vector instances, and position embedding from~\cite{devlin2018bert} to encode the relative positions of multiple points within a vector. These embedding are aggregated as the input of vector encoder, as shown in \cref{mee}.
In addition, we employ inter-instance and intra-instance attention mechanisms~\cite{maptrv2} to prioritize interactions within vectors over interactions between vectors, as depicted in the dashed box on the right side of \cref{mee}.
The \texttt{[VEC]} token in output serves as fused feature of rules and vectors, enabling the final prediction of their relationships through association head.

\subsection{Implementation}\label{method-overview}
We utilize VLE and MEE as backbones to integrate multiple modalities and address these two sub-tasks. The specific procedures are detailed as follows:

\paragraph{Rule Extraction from Traffic sign.}
To clarify the objectives of model, we first \emph{cluster symbols and texts into groups}. As shown in the upper part of \cref{overall}, the VLE is used to encode OCR results and images. By calculating the cosine similarity between \texttt{[STC]} tokens, different symbols and texts are clustered into groups. This process is supervised by contrastive loss during training.
Next, using grouped OCR results as text input and maintaining the VLE structure, we \emph{extract lane-level rules}. We employ multiple linear layers as understanding head for the \texttt{[CLS]} token to predict the corresponding value for each property of the rules. This facilitates to express all rules as $\{key:value\}$ pairs.
% Next, using grouped OCR results as text input and maintaining the VLE structure, we \emph{extract lane-level rules}. We employ a multi-classification head (understanding head) for the \texttt{[CLS]} token to predict the corresponding value for each property of the rules. This process allows us to express all rules inside a traffic sign as $\{key:value\}$ pairs.

\paragraph{Rule-Lane Correspondence Reasoning.} MEE is designed for vector encoding and interaction with rules. Each formatted rule is mapped to an embedding through MLP and fused with vector features in the fusion encoder, as shown in the lower part of \cref{overall}. We add a binary classification head after each \texttt{[VEC]} token to determine the relationship between the current centerline and rule. Details can be found in supplement.

\subsection{Experiment}\label{exp53}
\paragraph{Setups.} The dataset is split into $train$ and $test$ sets in the ratio of $9:1$. $L=6$ in VLE and $M=2, N=2$ in MEE. Input images are resized to $ 256 \times 256 $ and the feature dimension is $768$ with consistent $12$ attention heads. We initialize VLE with pre-trained weights of DeiT~\cite{deit} and BERT~\cite{devlin2018bert} while MEE is trained from scratch. The training procedure runs 50 and 120 epochs for VLE and MEE, respectively. More details can be found in the supplement. 

\paragraph{Results.} 
We use a heuristic method based on OCR character matching and nearest lane matching as the modular baseline. Meanwhile we make modifications to ALBEF~\cite{ALBEF} and BERT~\cite{devlin2018bert} to adapt them to our task.
As shown in \cref{exp_result}, 
the heuristic method performs poorly in both R.E. and C.R. tasks, underscoring the challenging nature of the MapDR dataset.
ALBEF-BERT method significantly enhances performance in the R.E. sub-task however lacking adequate adaptation to vector information hampers its performance on the C.R..
\cref{ablation-vle} indicates the attention mechanisms significantly improve $R_{R.E.}$, while layout of text brings marginal improvement. 
For C.R., the attention mechanisms is the critical part that promotes the effectiveness of MEE.
Instance embedding slightly improves $P_{C.R.}$ and $R_{C.R.}$, while type embedding significantly enhances both, indicating that vector types help the model establish rule-lane correspondence.
The separate evaluation results of all lane types and the strategies of heuristic method can be found in the supplement.

\begin{figure*}[t]
  \centering
  \includegraphics[width=0.9\textwidth]{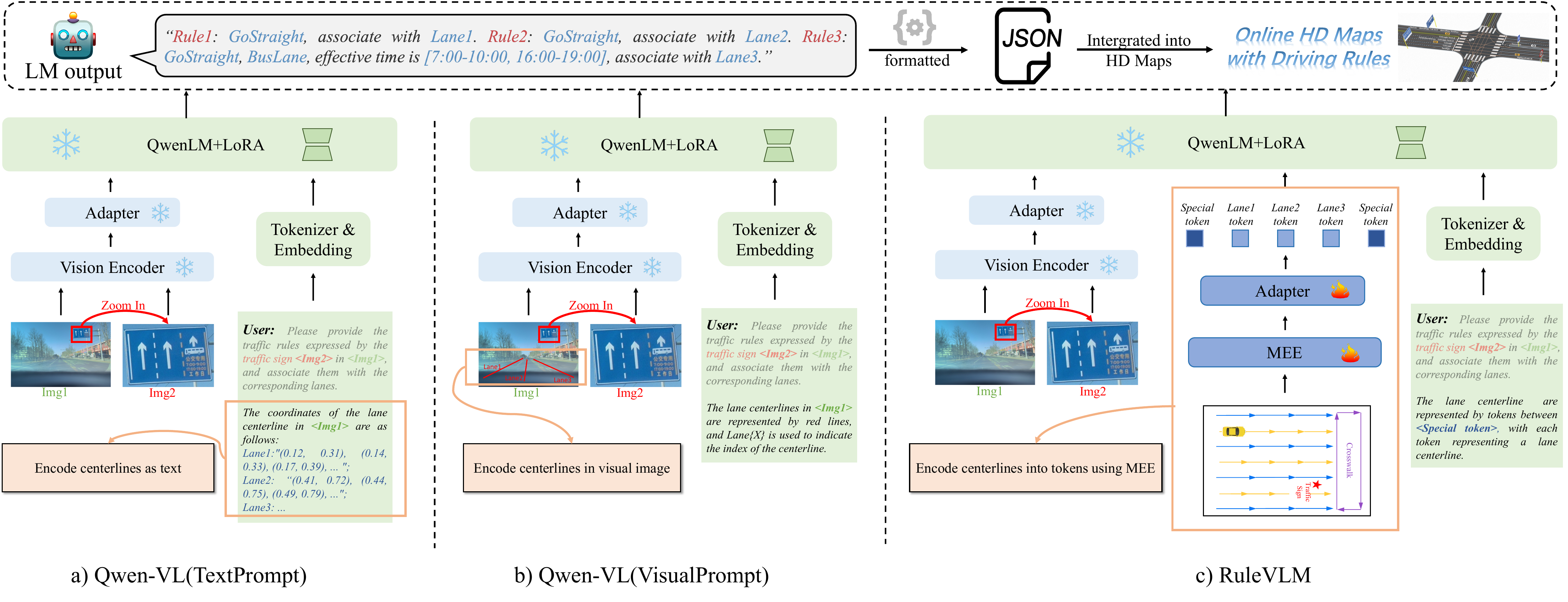}
  \caption{\textbf{Overview of end-to-end approaches.} Based on different vectorized HD maps encoding methods, approaches can be categorized into three types.}
  \label{qwendr}
\end{figure*}

\begin{table*}[t]
  \centering
\caption{\textbf{Evaluation of the overall task.} 
The heuristic method and the Qwen-VL(TextPrompt) method serve as the baselines for the modular and end-to-end approach, respectively. "$-$" denotes end-to-end approach is not suitable to independent evaluations of C.R. because these approaches do not utilize ground truth of rules for correspondence reasoning independently.}
  \resizebox{\textwidth}{!}{%
  \begin{tabular}{llccccccc}
  \toprule
    \multirow{2}{*}{\textbf{Model}} & \multirow{2}{*}{\textbf{Type}} &\multicolumn{2}{c}{\textbf{R.E.}}  &\multicolumn{2}{c}{\textbf{C.R.}}  &\multicolumn{3}{c}{\textbf{Overall}} \\ 
    \cmidrule(r){3-4} \cmidrule(r){5-6} \cmidrule(r){7-9}
    & & \bm{$P_{R.E.}(\textbf{\%})$} & \bm{$R_{R.E.}(\textbf{\%})$} & \bm{$P_{C.R.}(\textbf{\%})$} & \bm{$R_{C.R.}(\textbf{\%})$} & \bm{$ P_{all}(\textbf{\%})$} & \bm{$\bm R_{all}(\textbf{\%})$} & \bm{$ F1~Score $}  \\
    \midrule
    Heuristic &  \multirow{3}{*}{Modular} & 18.01  &  11.51  & 33.05 & 17.99 & 5.01 & 2.73 & 0.035  \\ 
    ALBEF-BERT & & 75.78  &  57.56  & 4.14 & 17.25 & 0.24  & 0.78 & 0.003  \\ 
    \rowcolor{gray!15}
    \textbf{VLE-MEE } & & \textbf{76.67} & \textbf{74.54} & \textbf{78.05 } &  \textbf{82.16} &  \textbf{63.35}  & \textbf{67.37}  &  \textbf{0.653}  \\ 
    \midrule
    Qwen-VL(TextPrompt) & \multirow{3}{*}{End-to-End}  &  42.21 & 41.09 & $-$ &  $-$ &  8.39  & 8.17  & 0.083 \\ 
    Qwen-VL(VisualPrompt) & &  89.29 & 89.50 & $-$ &  $-$ &  39.14  & 39.23  &  0.392 \\ 
    \rowcolor{gray!15}
    \textbf{RuleVLM} & &  \textbf{89.28} & \textbf{89.44} & $-$ &  $-$ &  \textbf{64.16}  & \textbf{64.28}  &  \textbf{0.642} \\ 
    \bottomrule
    \end{tabular}
  }
  \label{exp_result}
\end{table*}

\begin{table}[t]
    \caption{\textbf{Evaluation of sub-tasks.} 
    % Left: Rule Extraction, Right: Correspondence Reasoning. 
    "Attn." indicates intra \& inter-instance attention mechanisms. "Layout" refers to the text layout applied in VLE.  "In.E." and "Ty.E." denotes instance and type embedding in MEE, respectively.}
    \centering
    % \begin{minipage}{\columnwidth}
        \centering
        \resizebox{0.75\linewidth}{!}{%
        \begin{tabular}{cccc}
            \toprule
            \multicolumn{2}{c}{\textbf{VLE}} & \multirow{2}{*}{$\bm{P_{R.E.}(\textbf{\%})}$} & \multirow{2}{*}{$\bm{R_{R.E.}(\textbf{\%})}$} \\ \cmidrule{1-2}
            \textbf{Attn.} & \textbf{Layout} &  &  \\
            \midrule
            \ding{55} & \ding{55} & 75.78 & 57.56  \\ 
            \ding{51} & \ding{55} & 76.86 & 71.75    \\ 
            \rowcolor{gray!15}
            \ding{51} & \ding{51} & \textbf{76.67} & \textbf{74.54}   \\ 
            \bottomrule
        \end{tabular}
        }
        \label{ablation-vle}
    % \end{minipage}
    % \hfill
    % \begin{minipage}{\columnwidth}
        \centering
        \resizebox{0.85\linewidth}{!}{%
        \begin{tabular}{ccccc}
            \toprule
            \multicolumn{3}{c}{\textbf{MEE}} & \multirow{2}{*}{$\bm{P_{C.R.}(\textbf{\%})}$} & \multirow{2}{*}{$\bm{R_{C.R.}(\textbf{\%})}$} \\ \cmidrule{1-3}
            \textbf{Attn.} & \textbf{In.E.} & \textbf{Ty.E.} & & \\
            \midrule
            \ding{55}  & \ding{55} & \ding{55} & 4.14  & 17.25        \\
            \ding{51}  & \ding{55} & \ding{55} & 68.91 & 71.39    \\ 
            \ding{51}  & \ding{51} & \ding{55} & 69.68 & 72.76    \\ 
            \rowcolor{gray!15}
            \ding{51}  & \ding{51} & \ding{51}   & \textbf{78.05}  & \textbf{82.16}   \\ 
            \bottomrule
        \end{tabular}
        }
        \label{ablation-mee}
    % \end{minipage}
    \label{tab:table}
\end{table}

\section{End-to-End Approach}\label{end2end_approach}
Recently, there has been notable progress in the development of Multimodal Large Language Models (MLLMs)~\cite{DBLP:gpt4, Qwen-VL, claude, geminiteam2024gemini}, demonstrating significant capabilities in processing multimodal data and acquiring implicit knowledge of modality alignment. Their large-scale pre-training endows them with exceptional understanding and generative abilities, thereby enabling the comprehensive execution of the task proposed by MapDR in an end-to-end manner.
This section aims to address the overall task using Qwen-VL~\cite{Qwen-VL} and to investigate three distinct vector encoding methods. Moreover, we introduce an innovative \textbf{RuleVLM} model based on MEE, designed to process inputs from three modalities (images, texts, and vectors) and to produce driving rules, which yield favorable results on MapDR. Further elaboration on these concepts is presented in subsequent sections.

\subsection{Architecture}
As shown in \cref{qwendr}, we classify MLLM methods into three types: Qwen-VL(TextPrompt), Qwen-VL(VisualPrompt), and RuleVLM, based on different vector encoding methods. These methods utilize the framework of Qwen-VL~\cite{Qwen-VL}, with ViT~\cite{dosovitskiy2020vit} serving as the image encoder. Qwen-VL(TextPrompt) encodes centerline coordinates in the PV image as text, inputs them into QwenLM; Qwen-VL(VisualPrompt) visualizes the centerline and its index in the PV image and inputs it as Visual Prompt into QwenLM, while RuleVLM uses MEE with the cross-attention layer removed to encode vectorized centerline results and aligns it with LLM through an adapter. The goal of all these methods is to output a serialized set of driving rules in text form, decode it through a json decoder to restore formatted data, and then integrate it into HD maps for further applications.

\subsection{Implementation}
The JSON-formatted driving rules in MapDR are serialized into text and used as the answers in the QA format for training and evaluation.
For all methods based on QwenVL, the pre-trained weights of Qwen-VL-Chat (9.6B) are used for initialization. For method using MEE, the trained weights of MEE from the modular approach are utilized for its initialization.
During the supervised fine-tuning, we perform LoRA~\cite{lora} fine-tuning of the QwenLM, while keeping the parameters of ViT frozen.
For RuleVLM, we employ a linear layer as the adapter between QwenLM and MEE, while the parameters of MEE and the adapter remains trainable. 
We use special tokens to separate the centerline tokens extracted by MEE from the visual and text tokens. Additionally, to prevent overfitting, we shuffle the order of centerlines in the local vectorized map during training.

\subsection{Experiment.}
\paragraph{Setups.} 
All experiments are conducted on QA formatted MapDR dataset with 20 epochs of training on NVIDIA RTX A6000. The dimension of hidden states remains $4096$ while $r$ and $alpha$ of LoRA configuration are set $64$ and $16$ respectively. Meanwhile training batch size is 6 and the max training token length remains 2048.

\paragraph{Results.} 
We conducted zero-shot evaluations on a subset of MapDR using existing MLLMs~\cite{DBLP:gpt4,geminiteam2024gemini,claude,Qwen-VL} as qualitative investigation, but we encounter challenges in achieving stable structured outputs, further details are provided in the supplement.
Additionally, as shown in the \cref{exp_result}, Qwen-VL(TextPrompt) method serves as a end-to-end baseline do not perform better on both the R.E. and overall evaluation. Textual representations of centerline coordinates are challenging for MLLMs to perform spatial reasoning and also resulted in excessive length of sequence, which affected $P_{R.E.}$ and $R_{R.E.}$. 
Qwen-VL(VisualPrompt) visualizes the centerline within the image, resulting in reduced text sequence length, and demonstrates notable enhancements in the R.E. task compared to the baseline. However, the overall metrics are suboptimal, suggesting that image-based spatial reasoning remains an exceedingly challenging endeavor.
The RuleVLM method that we propose introduces MEE for the independent encoding of vectorized HD maps information. This facilitates a more direct extraction of centerline features, resulting in significant enhancements in the overall metrics.

\section{Conclusion} \label{conclusion}

In this work, we defined the task of integrating traffic regulations into vectorized HD maps to facilitate the construction of the traffic regulation layer within online HD maps. We constructed the MapDR dataset, which includes over $10,000$ video clips, more than $400,000$ images, and at least $18,000$ driving rules, covering a total mileage of approximately $800Km$. Additionally, we proposed the modular approach MEE-VLE and the end-to-end approach RuleVLM, establishing effective baselines for further research.
We hope that MapDR can serve as a starting point, and that more researchers can contribute to expanding this dataset or proposing even better solutions.

\clearpage
\newpage
{
    \small
    \bibliographystyle{ieeenat_fullname}
    \bibliography{main}
}

% CVPR 2025 Paper Template; see https://github.com/cvpr-org/author-kit

% \documentclass[10pt,twocolumn,letterpaper]{article}

%%%%%%%%% PAPER TYPE  - PLEASE UPDATE FOR FINAL VERSION
% \usepackage{cvpr}              % To produce the CAMERA-READY version
% \usepackage[review]{cvpr}      % To produce the REVIEW version
% \usepackage[pagenumbers]{cvpr} % To force page numbers, e.g. for an arXiv version

% Import additional packages in the preamble file, before hyperref
% \input{preamble}

% It is strongly recommended to use hyperref, especially for the review version.
% hyperref with option pagebackref eases the reviewers' job.
% Please disable hyperref *only* if you encounter grave issues, 
% e.g. with the file validation for the camera-ready version.
%
% If you comment hyperref and then uncomment it, you should delete *.aux before re-running LaTeX.
% (Or just hit 'q' on the first LaTeX run, let it finish, and you should be clear).

\definecolor{cvprblue}{rgb}{0.21,0.49,0.74}
% 定义浅紫色和浅蓝色
\definecolor{lightpurple}{RGB}{200, 170, 180}
\definecolor{lightblue}{RGB}{140, 180, 170}

%%%%%%%%% PAPER ID  - PLEASE UPDATE
\def\confName{CVPR}
\def\confYear{2025}

% \title{Driving by the Rules: A Benchmark for Integrating Traffic Sign Regulations into Vectorized HD Map}
%%%%%%%%% AUTHORS - PLEASE UPDATE
% \begin{document}
\newpage
\maketitlesupplementary

\appendix
\colorlet{punct}{red!60!black}
\definecolor{background}{HTML}{EEEEEE}
\definecolor{delim}{RGB}{20,105,176}
\colorlet{numb}{magenta!60!black}
\lstdefinelanguage{json}{   
    backgroundcolor=\color{white},
    basicstyle=\fontsize{7.5pt}{8.5pt}\fontfamily{lmtt}\selectfont,
    columns=fullflexible,
    breaklines=true,
    captionpos=b,
    commentstyle=\fontsize{8pt}{9pt}\color{codeblue},
    keywordstyle=\fontsize{8pt}{9pt}\color{colorred},
    stringstyle=\fontsize{8pt}{9pt}\color{codeblue},
    frame=tb,
    otherkeywords = {self},
    % basicstyle=\ttfamily\scriptsize,
    % numbers=none,
    % showstringspaces=false,
    % breaklines=true,
    % frame=lines,
    % backgroundcolor=\color{background},
    % keywordstyle=\color{blue}, % 关键字颜色
    % stringstyle=\color{green}, % style of strings
    literate=
     *{0}{{{\color{numb}0}}}{1}
      {1}{{{\color{numb}1}}}{1}
      {2}{{{\color{numb}2}}}{1}
      {3}{{{\color{numb}3}}}{1}
      {4}{{{\color{numb}4}}}{1}
      {5}{{{\color{numb}5}}}{1}
      {6}{{{\color{numb}6}}}{1}
      {7}{{{\color{numb}7}}}{1}
      {8}{{{\color{numb}8}}}{1}
      {9}{{{\color{numb}9}}}{1}
      {:}{{{\color{punct}{:}}}}{1}
      {,}{{{\color{punct}{,}}}}{1}
      {\{}{{{\color{delim}{\{}}}}{1}
      {\}}{{{\color{delim}{\}}}}}{1}
      {[}{{{\color{delim}{[}}}}{1}
      {]}{{{\color{delim}{]}}}}{1},
}

% \clearpage
% \tableofcontents

\section{Appendix Overview}
Our appendix encompass author statements, licensing, dataset access, dataset analysis, and the implementation details of benchmark results to ensure reproducibility. Additionally, we offer dataset documentation in adherence to the Datasheet format~\cite{gebru2021datasheets}, which covers details such as data distribution, maintenance plan, composition, collection, and other pertinent information.

\section{Author Statement}
We bear all responsibilities for licensing, distributing, and maintaining our dataset.

\section{Licensing}
The proposed dataset MapDR is under the CC BY-NC-SA 4.0 license, while the evaluation code is under the Apache License 2.0.

\section{Datasheet}
\subsection{Motivation}
\paragraph{For what purpose was the dataset created?}
Autonomous driving not only requires attention to the vehicle's trajectory but also to traffic regulations. However, in the online-constructed vectorized HD maps, traffic regulations are often overlooked. Therefore, we propose this dataset to integrate lane-level regulations into the vectorized HD maps. These regulations can serve as navigation data for both human drivers and autonomous vehicles, and are crucial for driving behavior.

\subsection{Distribution}
\paragraph{Will the dataset be distributed to third parties outside of the entity (e.g., company, institution, organization) on behalf of which the dataset was created?}
Yes, the dataset is open to public.

\paragraph{How will the dataset be distributed (e.g., tarball on website, API, GitHub)?} 
% The dataset will be made public on \emph{Tianchi} or \emph{ModelScope}, while the evaluation code will be publicly released on \emph{GitHub}.
The dataset is available at \href{https://modelscope.cn/datasets/MIV-XJTU/MapDR}{https://modelscope.cn/datasets/MIV-XJTU/MapDR}. Code is available at \href{https://github.com/MIV-XJTU/MapDR}{https://github.com/MIV-XJTU/MapDR}.

\subsection{Maintenance}
\paragraph{Is there an erratum?}
No. We will make a statement if there is any error are found in the future, we will release errata on the main web page for the dataset.

\paragraph{Will the dataset be updated (e.g., to correct labeling errors, add new instances, delete instances)?}
Yes, the dataset will be updated as necessary to ensure accuracy, and announcements will be made accordingly. These updates will be posted on the dataset's webpage on \href{https://modelscope.cn/datasets/MIV-XJTU/MapDR}{https://modelscope.cn/datasets/MIV-XJTU/MapDR}.

\paragraph{Will older versions of the dataset continue to be supported/hosted/maintained?}
Yes, older versions of the dataset will continue to be maintained and hosted.

\subsection{Composition}
\paragraph{What do the instances that comprise the dataset represent?}
An instance of the dataset consists of three main parts: a video clip, basic information, and annotation. The video clip comprises at least $30$ continuous front-view image frames, with one frame captured every $2$ meters to ensure uniform spatial distribution. Basic information of each clip is presented in the form of a JSON file, including the locations of traffic sign, all lane vectors, camera intrinsic parameters, and the camera poses for each frame. Annotation is also organized in JSON format, containing multiple driving rules. Each rule consists of a set of properties in $\{key:value\}$ format, along with the index of each centerline associated. All coordinates are transferred to the ENU coordinate systems, consistent within each segment but distinct between segments. For safety and privacy reasons, reference points are not provided.

\paragraph{How many instances are there in total (of each type, if appropriate)?} 
MapDR is composed of $10,000$ newly collected traffic scenes with over $400,000$ front-view images, containing more than $18,000$ lane-level driving rules. 

\paragraph{Are relationships between individual instances made explicit?}
The frames in a single video clip are continuous in time with a uniform spatial distribution. All video clips are collected among different time periods with consistent capture equipment and vehicles

\paragraph{Are there recommended data splits (e.g., training, development/validation, testing)?}
We have partitioned the dataset into two distinct splits: training and testing.

\paragraph{Is the dataset self-contained, or does it link to or otherwise rely on external resources?}
MapDR is totally newly collected and self-contained. Front-view images are captured and all the vectors are generated by our vectorized algorithm. All driving rules and correspondence are manually annotated.

\subsection{Collection Process}
\paragraph{Who was involved in the data collection process (e.g., students, crowdworkers, contractors) and how were they compensated (e.g., how much were crowdworkers paid)?}
Based on our HD map annotation scheme and annotation team, we have provided high-quality annotations with the help of experienced annotators and multiple validation stages.

\subsection{Use}
\paragraph{What (other) tasks could the dataset be used for?}
MapDR focus on the primary task of integrating driving rules from traffic signs to vectorized HD maps, which can be divided into two distinct sub-tasks: rule extraction and rule-lane correspondence reasoning. Researchers can also adapt to other traffic scene tasks.

\section{Dataset Production} \label{appendix_data_production}
\subsection{Data Production Pipeline}
\paragraph{Data Collection.}
Search and Retrieval: We use our own database to locate the GPS coordinates of traffic signs, utilizing both text-based and image-based retrieval methods.
Route Planning: Our path planning algorithm is employed to design data collection routes. Vehicles equipped with data collection devices gather raw data, including images, camera parameters, and pose information, which are then uploaded to the cloud.
\paragraph{Data Processing: Vectorization.}
 In the cloud, BEV (Bird's Eye View) perception algorithms are applied to generate vectorized local HD maps. Key point detection and matching algorithms are used to recover the 3D positions of traffic signs.
\paragraph{Rule Extraction.}
For each set of multiple image frames containing traffic signs, the most representative frame is selected for rule extraction by annotators. Vectorized map results are provided for annotating rule-lane associations. All captured images and the projection of vectorized maps in these images are included as reference material to enhance annotation accuracy.

\subsection{Annotation Process}
\paragraph{Rule Identification.}
Annotators identify the number of rules on each traffic sign and group related text information corresponding to each rule.
\paragraph{Annotation Creation.}
A JSON file is created with eight properties that annotators fill based on their interpretation of the rules.
\paragraph{Vector Association.}
Each rule is associated with the vector ID corresponding to its location on the vectorized map. Unique IDs are assigned to all vectors.
\paragraph{Quality Assurance.}
Quality inspection procedures are implemented to ensure the accuracy of annotations. This includes a thorough review and rework process to correct any discrepancies.

\section{Analysis of MapDR}
\paragraph{Data\&Label Composition.}
MapDR is organized into video clips, with each clip focusing on a single traffic sign. The raw data and annotation are provided as JSON files. 
We provide the detailed JSON schema of both files.
\cref{data_schema} is the JSON schema of data file (data.json). 
% \cref{data} demonstrates the composition of raw data. 
An example is as shown in \cref{data_demo}. 
The 3D spatial location of the traffic sign is provided by $4$ points represented as \textit{traffic\_board\_pose}. Vectors and their types are also provided. Additionally, camera intrinsics and pose for each frame are provided to facilitate vector visualization. Note that all coordinates have been transferred to relative ENU coordinate systems which is consistent within a clip. 
\cref{label_schema} is the JSON schema of annotation file (label.json). 
An example is as shown in \cref{label_demo}. All pre-defined properties of driving rules are illustrated. The corresponding centerlines of each rule are annotated by the vector index. As mentioned in main submission, spatial location of the symbols and texts which represent the particular rules, referred to as semantic groups, is also provided. Researchers can optionally utilize this information.  

\begin{figure*}[t]
    \centering
    \includegraphics[width= \columnwidth]{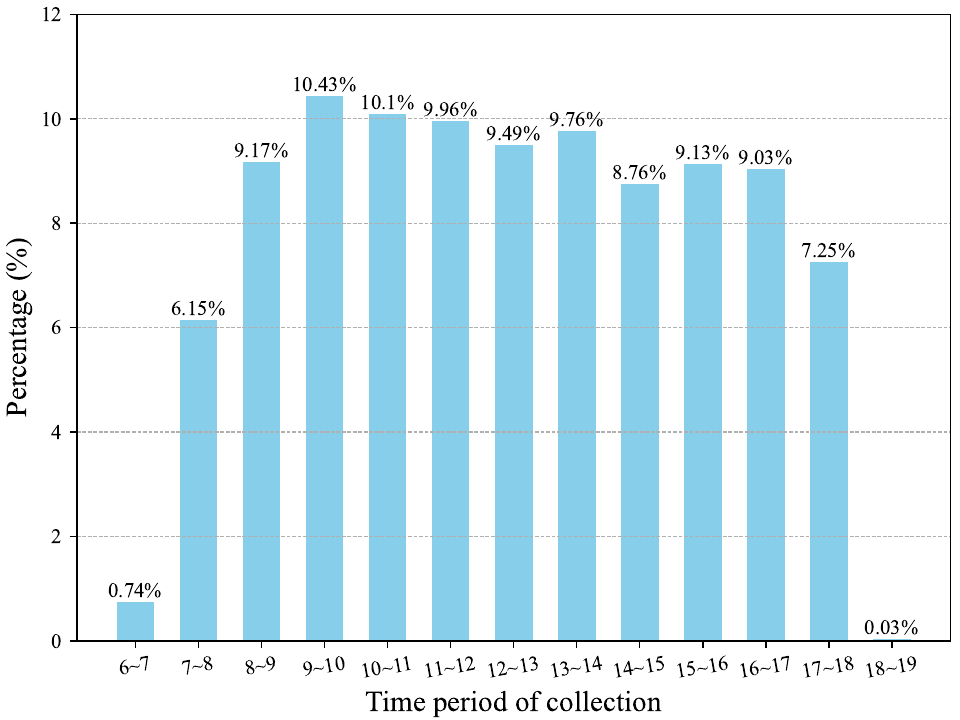} 
    \includegraphics[width= \columnwidth]{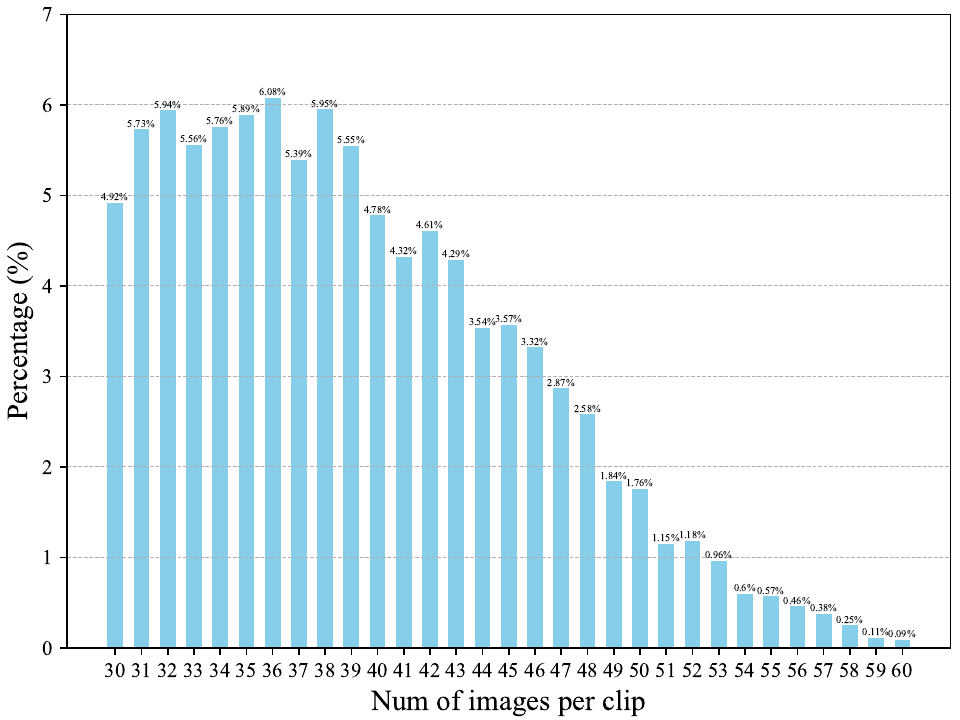}   
    \caption{\textbf{Metadata distribution of MapDR.}}
    \label{distribution}
\end{figure*}

\paragraph{Distribution of MapDR.}
\cref{distribution} illustrates the diverse metadata distribution in the MapDR dataset. 
Upper depicts the distribution of the time period for data collection, primarily from $07:00$ AM to $18:00$ PM, indicating that the dataset was mainly collected during daytime.
The lower displays the majority of clips containing between $30$ and $45$ frames.

\paragraph{Auxiliary Evaluation Results.} \label{appendix_eval_res_per_class}
We conducted separate evaluations on all traffic signs of different lane types in MapDR. As shown in \cref{eval_res_per_lanetype}, the results indicate that the prediction difficulty varies among different categories of traffic signs. 

\begin{table}[hb]
  \centering
    \caption{\textbf{Evaluation results of all traffic signs with different lane types in MapDR.} The results are all based on proposed modular method, and the split of dataset remains unchanged.}
    \centering
    \resizebox{\columnwidth}{!}{%
    \begin{tabular}{lcccc}
    \toprule
    \textbf{Metric}  & BusLane & DirectionLane&	EmergencyLane&VariableDirectionLane\\
    \midrule
    \textbf{$P_{R.E.}(\%)$} & 73.44 &78.44	&92.20	&71.42	 \\
    \textbf{$R_{R.E.}(\%)$} & 71.98	&77.36	&91.03	&57.14	 \\
    \textbf{$P_{C.R.}(\%)$} & 73.34	&82.12	&92.85	&71.42	 \\
    \textbf{$R_{C.R.}(\%)$} & 76.76	&87.03	&91.00	&85.71	\\
    \bottomrule
    \end{tabular}
    }
    \resizebox{\columnwidth}{!}{%
    \begin{tabular}{lccccc}
    \toprule
    \textbf{Metric}  &NonMotorizedLane&	VehicleLane &TidalFlowLane	&MultiLane	&SpeedLimitedLane \\
    \midrule
    \textbf{$P_{R.E.}(\%)$} &80.00	&88.88	&0	&82.09	&60.34 \\
    \textbf{$R_{R.E.}(\%)$} &72.00	&74.41	&0	&82.56	&53.85 \\
    \textbf{$P_{C.R.}(\%)$} &85.41	&61.90	&0	&81.33	&88.15 \\
    \textbf{$R_{C.R.}(\%)$} &83.67	&72.22	&0	&83.94	&97.10 \\
    \bottomrule
    \end{tabular}
    }
    \label{eval_res_per_lanetype}
\end{table}

\paragraph{Potential negative societal impacts.}
To minimize negative societal impact, we have applied obfuscation techniques to license plate numbers, facial features, and other personally identifiable information in our dataset. Additionally, sensitive geographical locations have been excluded, and coordinates in the ENU coordinate system have been provided without reference points to safeguard privacy. 
However, considering the potential inaccuracies and deviation of data distribution, the model may have misinterpretations and biases during the learning process. If such models are used on public roads, it could pose safety issues. Therefore, we recommend thorough testing of models before deploying to any autonomous driving system.

\section{Visualization of MapDR} \label{appendix_visualization}
\begin{figure*}[h]
    \includegraphics[width = \textwidth]{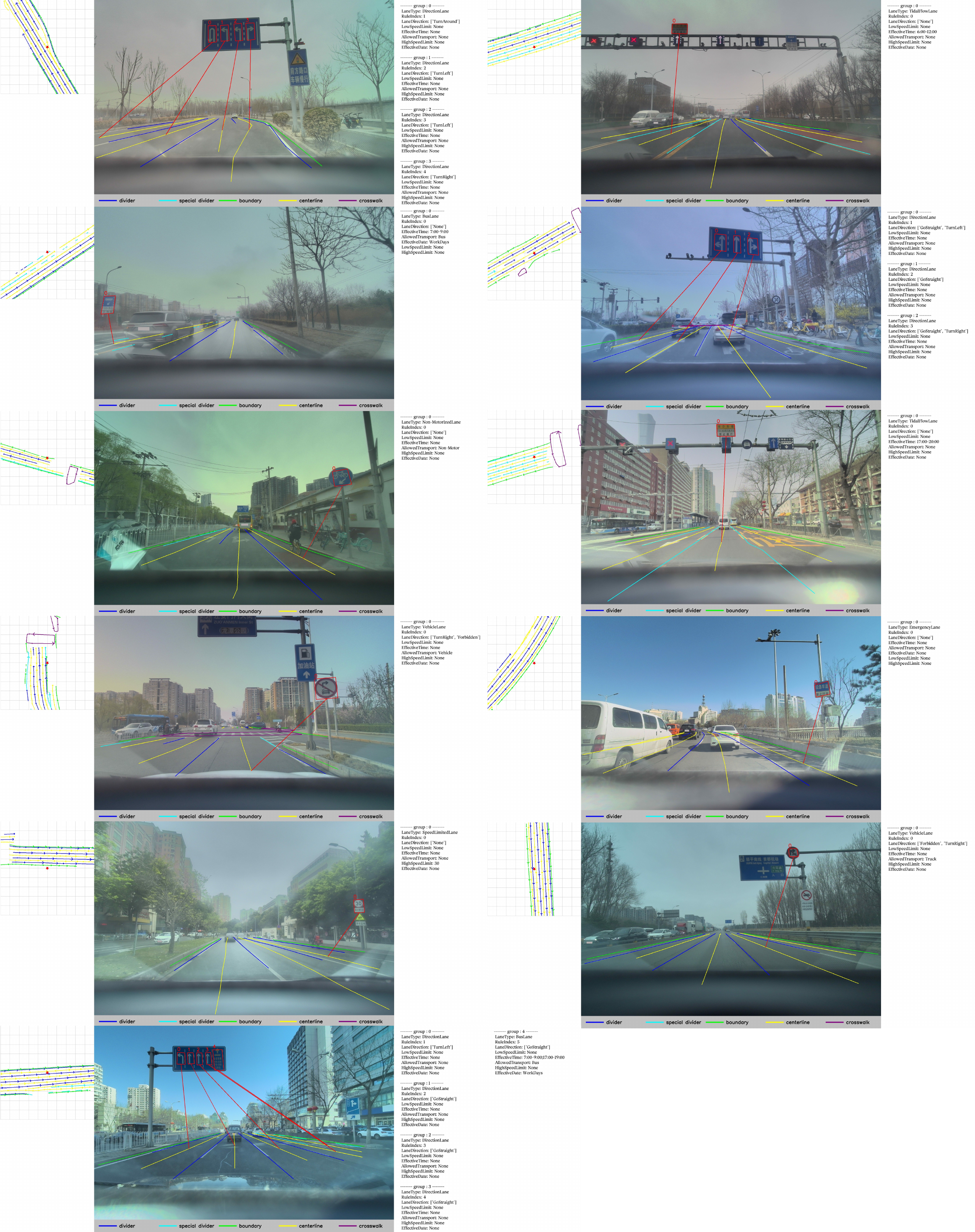}
    \caption{Visualization of MapDR.}
    \label{supple_vis}
\end{figure*}

\begin{figure*}[h]
    \includegraphics[width = \textwidth]{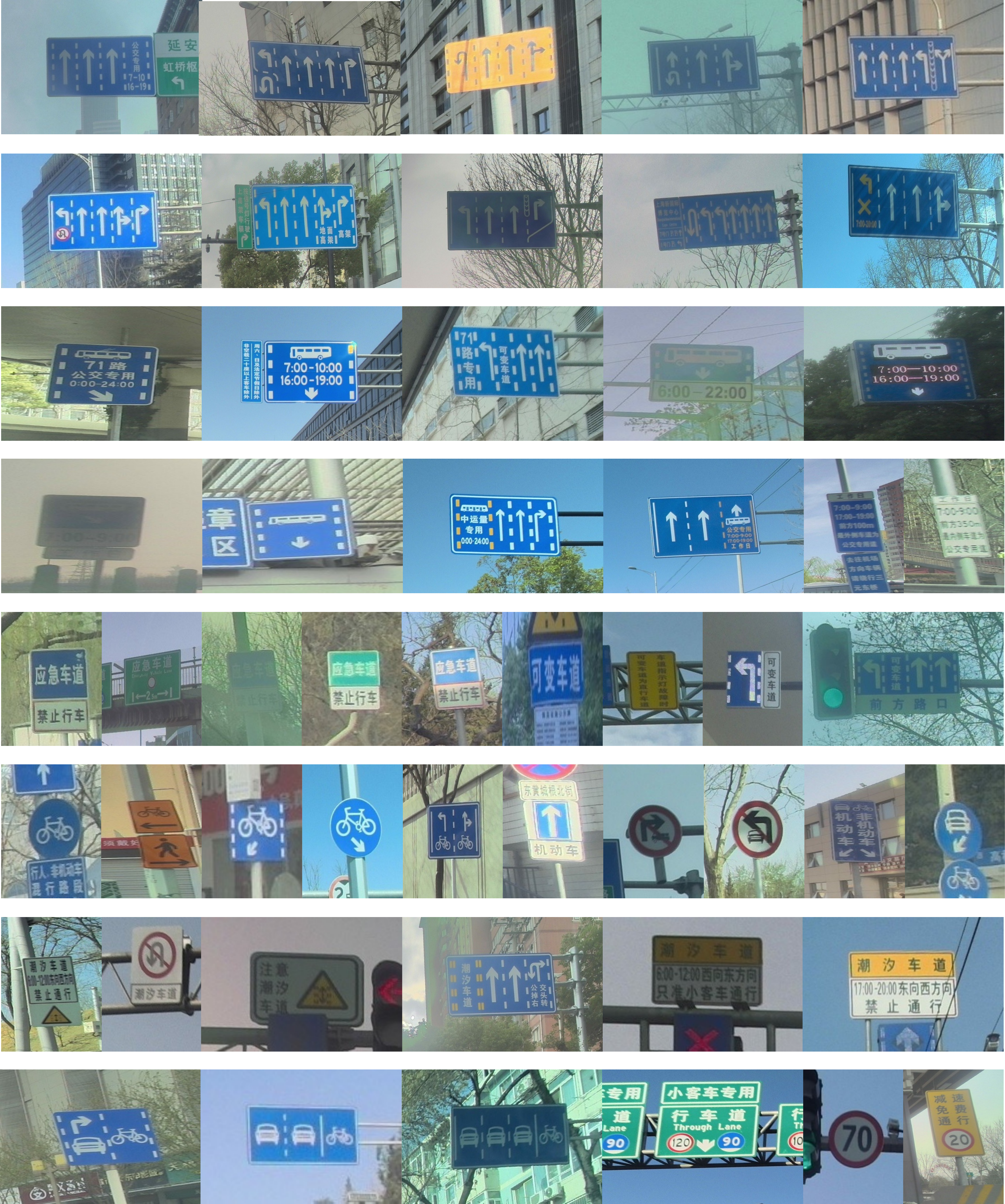}
    \caption{Visualization of traffic signs.}
    \label{traffic_sign_vis}
\end{figure*}

\cref{supple_vis} visualizes driving rules for different lane types in the dataset, including BEV and front-view images, as well as formatted driving rules. The red pentagram in the BEV image marks the position of the traffic sign. The front-view image displays the lane vectors and manually annotated semantic groups, with driving rules organized as sets of $\{key:value\}$ pairs.
\cref{traffic_sign_vis} shows diverse types of traffic signs collected at different times, locations, and weather conditions, demonstrating rich inter-class differences and intra-class diversity, highlighting the complexity of the MapDR dataset.

\section{Example for Evaluation Metric} \label{appendix_eval_example}
\begin{figure}[hb]
    \centering
    \includegraphics[width=\columnwidth]{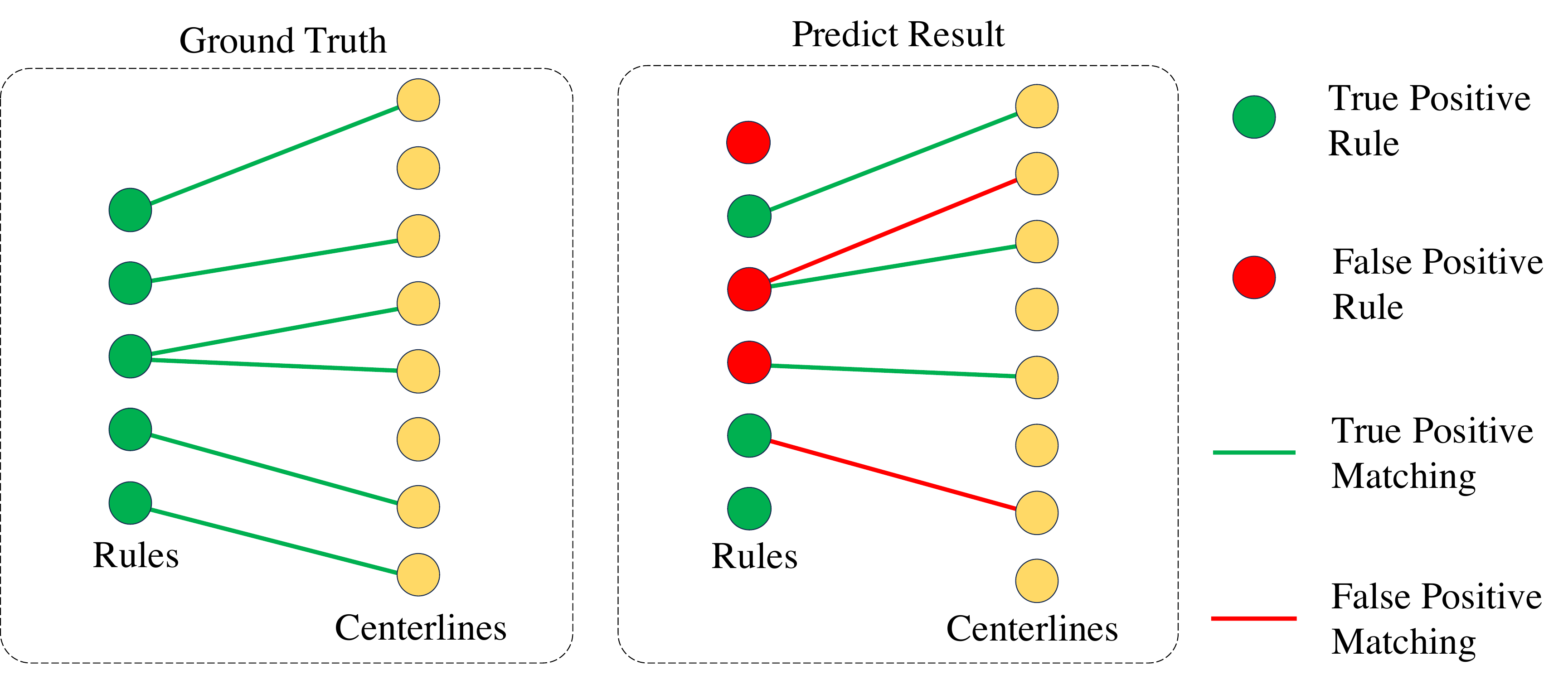}
    \caption{\textbf{Illustration example for Evaluation Metrics.}}
    \label{metric_example}
\end{figure}

We provide an example of metric calculation as \cref{metric_example} shown, illustrating the evaluation process. Given the ground truth $G$ with $5$ rule nodes and $8$ centerline nodes while $6$ edges between them, we assume that the algorithm has predicted $\hat{G}$ with $6$ rules and $5$ edges, the metric calculation process is detailed as below.

First, for the \textbf{Rule Extraction from Traffic Sign} sub-task, the ground truth has $5$ rules, while the algorithm predicted $6$ rules, of which $3$ are correct (\textcolor{green}{green circles}) and $3$ are incorrect (\textcolor{red}{red circles}). Then the precision ($P_{R.E.}$) and recall ($R_{R.E.}$) are calculated as \cref{eval_demo_1}:

\begin{equation}
\begin{aligned}
    &P_{R.E.} = \frac{| \hat{R} \cap R |}{|\hat{R}|} =  \frac{3}{6} 
  \qquad R_{R.E.} = \frac{|\hat{R} \cap R | }{| R | } =  \frac{3}{5} \\
  \label{eval_demo_1}
\end{aligned}
\end{equation}

Next, for the \textbf{Rule-Lane Correspondence Reasoning} task, there are $6$ association results in the ground truth, but the algorithm predicted $5$, with $3$ being correct (\textcolor{green}{green lines}) and $2$ being incorrect (\textcolor{red}{red lines}). Then, the precision ($P_{C.R.}$) and recall ($R_{C.R.}$) are calculated as \cref{eval_demo_2}:

\begin{equation}
\begin{aligned}
  &P_{C.R.} = \frac{| \hat{E} \cap E |}{|\hat{E}|} =  \frac{3}{5}   
  \qquad R_{C.R.} = \frac{| \hat{E} \cap E | }{| E |} =  \frac{3}{6} \\
  \label{eval_demo_2}
\end{aligned}
\end{equation}

Finally, considering the entire task, in the ground truth, a total of $6$ lanes are assigned driving rules. The model predicted driving rules for $5$ lanes, with correct predictions for both the association relationship and driving rules for only $1$ lane. Therefore, the precision ($P_{all}$) and recall ($R_{all}$) for the entire task are calculated as \cref{eval_demo_3}:

\begin{equation}
\begin{aligned}
  &P_{all} = \frac{| \hat{G^{s}} \cap G^{s} |}{|\hat{G^{s}}|} =  \frac{1}{5}  
  \qquad R_{all} = \frac{|\hat{G^{s}} \cap G^{s}|}{|G^{s}| }=  \frac{1}{6} \\
  \label{eval_demo_3}
\end{aligned}
\end{equation}

\section{Implementation Details} \label{appendix_exp_setup}

All experiments utilizing the modular approach are conducted on 8 NVIDIA V100 16G GPUs, whereas the end-to-end approach experiments are performed on 8 NVIDIA RTX A6000 48G GPUs.
We utilize pre-trained weights of DeiT~\cite{deit} and 
BERT~\cite{devlin2018bert} to initialize the modular model in our experiments. 
Assets of DeiT and BERT are licensed under the Apache-2.0 license. 
Pre-trained weights of Qwen-VL-Chat~\cite{Qwen-VL} are employed to initialize the end-to-end model and the weights are under Tongyi Qianwen license.
Additionally, we have adopted ALBEF~\cite{ALBEF} and Qwen-VL as our code base, which are available under the BSD 3-Clause and Tongyi Qianwen license respectively.

\subsection{Heuristic approach}
We design the heuristic method based on OCR character matching and nearest lane association. Specifically, we first perform OCR detection for the sign images then predict the values corresponding to different properties in the driving rule based on the presence of specific text or symbols in the OCR detection results. 
For example, if there is a "bus" symbol in the OCR result the 
\textit{"LaneType"} property will be predicted as \textit{"BusLane"}, meanwhile the \textit{"AllowdTransport"} property will be predicted as \textit{"Bus"}. 
If a text line contains purely numeric text similar to time or speed limits, its format is used to determine whether it represents the value of \textit{"Effective Time"}, \textit{"HighSpeedLimit"} and \textit{"LowSpeedLimit"}.
For C.R., we calculate the shortest distance between each centerline in the local vectorized HD map and the sign coordinates, selecting the nearest distance as the corresponding centerline associated with the rule. 

The experimet result of heuristic method indicates that relying solely on OCR and heuristics is insufficient for this complex task, which requires a more sophisticated approach integrating image features, OCR results, and layout analysis. We agree that while this method offers some insight, however it lacks long-term research value.

\subsection{Vision-Language Encoder (VLE)}
\paragraph{Hyperparameters and Configurations.}
We conduct $lr=1e-4 $, $warmup\_lr=1e-5$, $decay\_rate=1$, $weight\_decay=0.02$, $embedding\_dim=768$, $momentum =0.995$, $alpha =0.4$, $attention\_heads=12$, and $batch\_size =32$ for all experiments. 
We initialize vision encoder with pre-trained weight of DeiT~\cite{deit}, text encoder and fusion encoder with the first $6$ layers and last $6$ layers of BERT~\cite{devlin2018bert}, respectively. The fine-tuning epoch is set to $50$. Input image is resized to $256 \times 256$. The maximum number of tokens for input in the text encoder is $1000$. \emph{RandomAugment} is used, with hyperparameters $N=2$, $M=7$, and it includes the following data augmentations: \emph{"Identity", "AutoContrast", "Equalize", "Brightness", "Sharpness"}.

\paragraph{Clustering head.}
We calculate the cosine similarity between the \texttt{[STC]} tokens to determine if they represent the same rule. The training procedure is supervised by \emph{Contrastive Loss}. The positive margin is set to $0.7$, and the negative margin is set to $0.3$.
 % $\mathcal{L}_{contrastive}$

\paragraph{Understanding head.}
For properties in each rule, we prefer to classify their value into pre-defined classes. Specifically, for \textit{"RuleIndex"}, \textit{"LaneType"}, \textit{"AllowedTransport"}, \textit{"EffectiveDate"} we employ linear layer to perform classification with \emph{Cross-Entropy Loss}.
For \textit{"LaneDirection"}, this property is predicted by a multi-label classification that direction is defined as a combination of multi-choice from [\textit{"None"},\textit{"Forbidden"},\textit{"GoStraight"},\textit{"TurnLeft"},\textit{"TurnRight"}, \\
\textit{"TurnAround"}]. 
The training loss is \emph{Binary Cross-Entropy Loss}.
Additionally, properties of \textit{"EffectiveTime"}, \textit{"LowSpeedLimit"} and \textit{"HighSpeedLimit"} are formed as \textit{string}. In practice, we classify the \texttt{[STC]} token to determine whether the OCR text is time or speed and use the original OCR text as the predicted value of these three properties. 

\subsection{Map Element Encoder (MEE)}
\paragraph{Hyperparameters and Configurations.}
We conduct $lr=1e-4 $, $warmup\_lr=1e-5$, $decay\_rate=1$, $weight\_decay=0.02$, $embedding\_dim=768$, $momentum =0.995$, $alpha =0.4$, $attention\_heads=12$, and $batch\_size =48$ for all experiments. 
We train MEE from scratch, the training epoch is set to $120$. The maximum number of tokens for input in the vector encoder is $1000$. 
The formatted rule is mapped to a $768$-dimensional vector by an MLP. Specifically, each property in the rule is mapped to a $768$-dimensional vector (except for \textit{"EffectiveTime"}, \textit{"LowSpeedLimit"} and \textit{"HighSpeedLimit"}), and the position of the traffic sign is also mapped to a $768$-dimensional vector through a position encoding method (as described in the main submission), and finally, all these vectors are added together to obtain the final feature of the rule.
In MEE, there are a total of four types of embeddings: vector embedding, position embedding, type Embedding, and instance embedding. The encoding method for vector embedding and position Embedding is detailed in the main submission. For type embedding, as there are $5$ types in total, we initialize it using $nn.Embedding$, with the hyperparameters $num\_embeddings=5$ and $embedding\_dim=768$. Similarly, we also use $nn.Embedding$ to initialize the instance embedding, with the $num\_embeddings=120$ and $embedding\_dim=768$, meaning it can support a maximum of $120$ vectors. It is important to note that since the instance embedding is only used to distinguish different vectors, we shuffle the order of these embeddings at each iteration. After the multimodal fusion encoder of MEE, we further incorporate an $nn.Linear$ to map the $768$-dimensional features to $256$, which is then connected to the association head.

\paragraph{Association head.}
We perform binary classification on \texttt{[VEC]} tokens to determine whether the vector is corresponding to the input rule. The training procedure is supervised with \emph{Binary Cross-Entropy Loss}.
% $\mathcal{L}_{BinaryCrossEntropy}$. 

\subsection{Analysis of Evaluation Error}
\begin{figure}[t]
    \centering
    \includegraphics[width = \columnwidth]{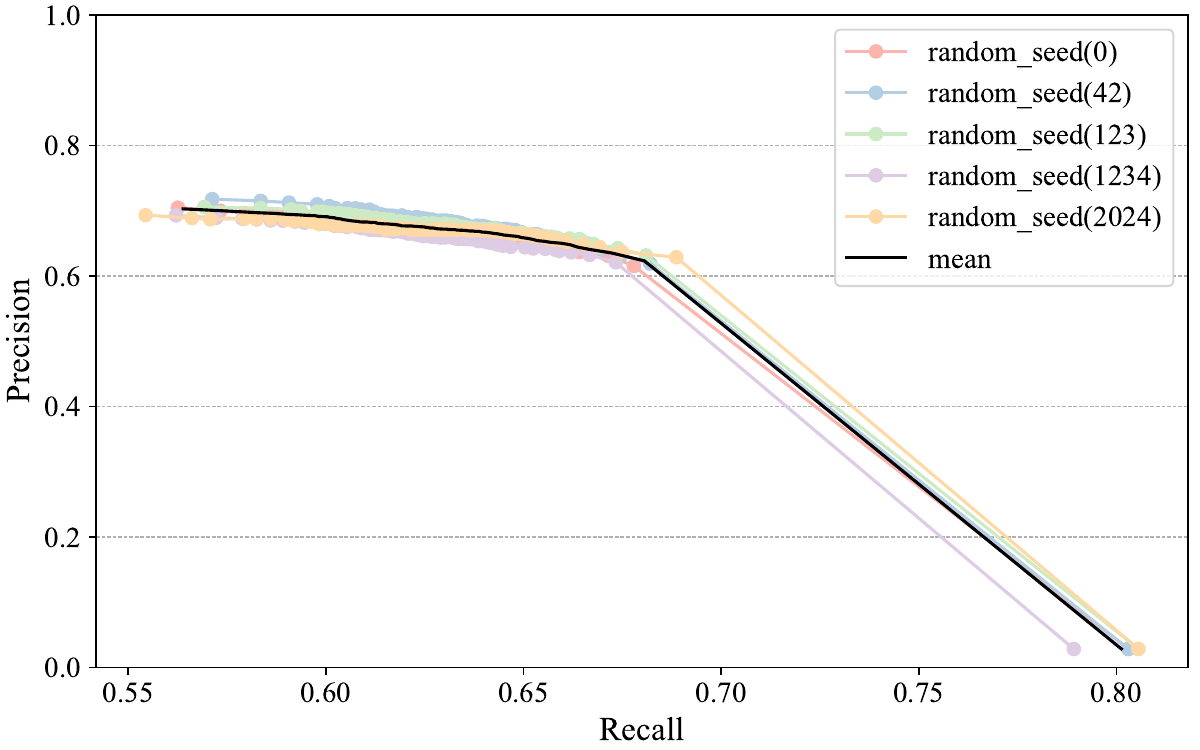}
    \caption{Overall P-R curves with various random seeds.}
    \label{error}
\end{figure}

We conduct multiple experiments on proposed modular approach with various random seed, and the experimental results are shown in \cref{error}. 
We repeated all experiments $5$ times with various seed which are depicted in different colors. We uniformly sampled $100$ points within the range of $0$ to $1$ as the binary classification threshold for association head in correspondence reasoning procedure, and then calculate the $P_{all}$ and $R_{all}$ for each threshold. The mean fitted line is shown in black, demonstrating the stability of our method. Specifically, we calculated the standard deviation of all evaluation metrics at a fixed threshold among different random seeds. For rule extraction sub-task, the standard deviation of $P_{R.E.}$ and $R_{R.E.}$ are $0.32$ and $0.38$. In the rule-lane correspondence reasoning sub-task the standard deviations are $0.07$ and $0.38$ for $P_{C.R.}$ and $R_{C.R.}$. Overall, the standard deviations of  $P_{all}$, $R_{all}$ and $AP$ are $0.18$ $0.10$ and $1.07$, respectively.

\section{Qualitative results of MLLM} \label{appendix_mllm}
We qualitatively evaluated the zero-shot performance of existing MLLMs on the two subtasks of \textbf{Rule Extraction} and \textbf{Correspondence Reasoning} using a subset of MapDR, which consists of $20$ randomly sampled examples for traffic signs among all lane types, totaling $180$ cases. 
Annotators subjectively assessed the correctness of MLLM outputs. Since MLLMs cannot provide confidence scores for their predictions, we could not use a threshold to calculate precision and recall metrics. 
Therefore, we evaluated $Acc_{R.E.}$ and $Acc_{C.R.}$ denotes the accuracy of R.E. and C.R. on the board-level. 
The evaluation reflect whether the model can interpret all the rules within a traffic sign and associated with correct centerlines, as shown in \cref{mllm_res}. 
% Therefore, we evaluated accuracy, specifically $Acc_{R.E.} = \frac{|\hat{R} \cap R | }{| R | }$ and $Acc_{C.R.} = \frac{|\hat{E} \cap E | }{| E | }$, as shown in Table \ref{mllm_res}.

\begin{table}[h]
  \centering
    \caption{\textbf{Zero-shot accuracy on the subset of MapDR.} MLLMs are subjectively evaluated by annotators, so the results only approximately reflect their capacity.}
    \centering
    \begin{tabular}{l cc}
    \toprule
    \textbf{Model}  & \textbf{$Acc_{R.E.}(\%)$}  & \textbf{$Acc_{C.R.}(\%)$} \\
    \midrule
    Qwen-VL Max~\cite{Qwen-VL}    & 44.4  & 20.6 \\
    Gemini Pro~\cite{geminiteam2024gemini} & 31.1  & 6.1  \\
    Claude3 Opus~\cite{claude}   & 4.4   & 1.1 \\
    GPT-4V~\cite{DBLP:gpt4}    & 3.3   & 1.7 \\
    % \rowcolor{gray!25}
    % Ours       & 65.15 & 78.84  \\ 
    \bottomrule
    \end{tabular}
    \label{mllm_res}
\end{table}

All existing MLLMs are evaluated without SFT, clearing former memories before each prompt to avoid contextual influence. This experiment primarily aims to qualitatively analyze the zero-shot capacity of MLLMs in traffic scene understanding, rather than a rigorous quantitative comparison. 
Overall, the results highlight the necessity of this task and dataset.

\begin{figure*}[h]
    \centering
    \includegraphics[width = 0.95\textwidth]{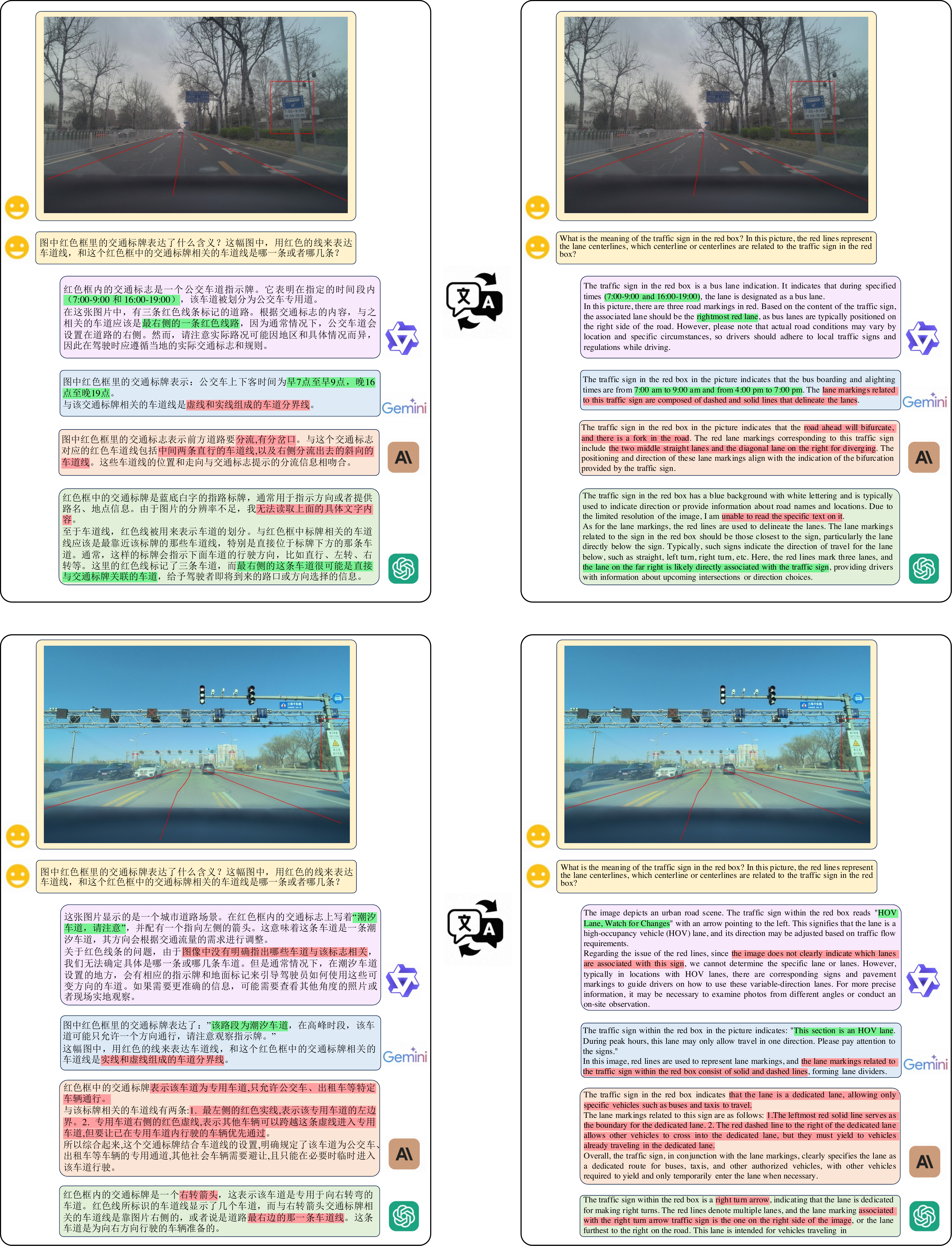}
    \caption{Prompts and answers for MLLMs.}
    \label{mllm_demo_vis}
\end{figure*}

As all the traffic signs and rules are from China, described in Chinese, we utilized a Chinese prompt. In \cref{mllm_demo_vis}, we present our input, including the image and prompt, along with the results generated by MLLMs. Our prompt can be translated as: \emph{"What is the meaning of the traffic sign in the red box? In this picture, the red lines represent the lane centerlines, which centerline or centerlines are related to the traffic sign in the red box?"}. The use of a Chinese prompt may also contribute to Qwen-VL's better performance, as it originates from Alibaba, a Chinese company, and its training process involved more Chinese text compared to other models~\cite{Qwen-VL}.

Additionally, we referenced~\cite{red-circle} to mark the red boxes and red lines in the images as visual prompts for the signs of interest and the centerlines of the lanes, which is convenient but may not be the most effective method and may also limit the performance of MLLMs. 
Furthermore, according to~\cite{llm-ocr}, we can learn that apart from the Qwen-VL model, other models such as GPT-4V have weak capabilities in Chinese OCR, so this possibly limit their cognitive performance. 
Overall, despite MLLMs' zero-shot performance not achieving remarkable results, they possess significant potential. We believe that with further prompt optimization, the implementation of SFT, and other methods, larger models will undoubtedly achieve improved results in the future.

\begin{figure*}[h]
\begin{lstlisting}[language=json, caption= Json schema of data file., label=data_schema]
json_schema =  {
    "$schema": "http://json-schema.org/draft-07/schema#",
    "type": "object",
    "properties": {
        "traffic_board_pose": {
            "type": "array",
            "minItems": 4,        
            "maxItems": 4,        
            "items": {
                "type": "array",
                "minItems": 3,    
                "maxItems": 3,    
                "items": {
                    "type": "number" }}
        },
        "vector": {
            "type": "object",
            "additionalProperties": { 
                "type": "object",
                "properties": {
                    "type": {
                        "type": "string",
                        "enum": ["0", "1", "2", "3", "4"] 
                    },
                    "vec_geo": {
                        "type": "array",
                        "items": {
                            "type": "array",
                            "minItems": 3,
                            "maxItems": 3,
                            "items": {
                                "type": "number"}}
                    }
                },
                "required": ["type", "vec_geo"],
                "additionalProperties": false}  
        },
        "camera_intrinsic_matrix": {
            "type": "array",
            "minItems": 3,  
            "maxItems": 3,
            "items": {
                "type": "array",
                "minItems": 3,  
                "maxItems": 3,
                "items": {
                    "type": "number"}}
        },
        "camera_pose": {
            "type": "object",
            "additionalProperties": { 
                "type": "object",
                "properties": {
                    "tvec_enu": {
                        "type": "array",
                        "minItems": 3,
                        "maxItems": 3,
                        "items": {
                            "type": "number"}},
                    "rvec_enu": {
                        "type": "array",
                        "minItems": 4,
                        "maxItems": 4,
                        "items": {
                            "type": "number"}}
                },
                "additionalProperties": false}
        }
    },
    "required": ["traffic_board_pose", "vector", "camera_intrinsic_matrix", "camera_pose"],
    "additionalProperties": false
}
\end{lstlisting}
\end{figure*}

\begin{figure*}[h]
\begin{lstlisting}[language=json, caption=Example of data file., label=data_demo]
{
    "traffic_board_pose": [
        [6250.741478919514, -23002.897461687568, -51.60124124214053],
        [6250.767766343895, -23002.852551855587, -53.601367057301104],
        [6247.90629957122, -23005.522309921853, -53.698920409195125],
        [6247.880012146425, -23005.5672197543, -51.69879459403455]
    ],
    "vector": {
        "0": {
            "type": "2",
            "vec_geo": [
                [6222.740794670596, -22977.551953653423, -59.28851334284991],
                [6224.65054626556, -22979.753116989126, -59.31985123641789],
                [6229.777790947785, -22985.886256590424, -59.40054347272962],
                [6237.236963539255, -22995.08138003234, -59.51233040448278],
                [6242.709547414123, -23002.134314719562, -59.58363144751638],
                [6247.894389983971, -23008.135111707456, -59.648408086039126],
                [6253.242476279292, -23014.058069147195, -59.700414426624775],
                [6258.56982873722, -23020.026259167204, -59.72872495371848]
            ]
        }
    },
    "camera_intrinsic_matrix": [
        [904.9299114165748, 0.0, 949.2163397703193],
        [0.0, 904.9866120329268, 623.7475554790544],
        [0.0, 0.0, 1.0]
    ],
    "camera_pose": {
        "1710907374739989000": {
            "tvec_enu": [6217.6643413086995, -22963.182929283157, -57.714795432053506],
            "rvec_enu": [-0.2097012215148481, 0.6478309996572192, -0.6804515437189796, 0.2707879063036554]
        }
    }
}
\end{lstlisting}
\end{figure*}

% \onecolumn
% \lstinputlisting[language=json, breaklines=true, caption=Json schema of label file., label=label_schema]{json/label_schema.json}
% \twocolumn

\begin{figure*}[h]
\begin{lstlisting}[language=json, caption=Json schema of label file., label=label_schema]
json_schema = {
    "$schema": "http://json-schema.org/draft-07/schema#",
    "type": "object",
    "additionalProperties": {
        "type": "object",
        "properties": {
            "attr_info": {
                "type": "object",
                "properties": {
                    "LaneType": {
                        "type": "string",
                        "enum": ["DirectionLane","BusLane","EmergencyLane","MultiLane","Non-MotorizedLane","SpeedLimitedLane","TidalFlowLane","VariableDirectionLane","VehicleLane"]},
                    "RuleIndex": {
                        "type": "string",
                        "enum": ["None", "1", "2", "3", "4", "5", "6", "7", "8", "9", "10"]},
                    "LaneDirection": {
                        "type": "array",
                        "items": {
                            "type": "string",
                            "enum": ["GoStraight","TurnLeft","TurnRight","TurnAround","Forbidden","None"]},
                        "minItems": 1,
                        "maxItems": 5},
                    "AllowedTransport": {
                        "type": "string",
                        "enum": ["None", "Vehicle", "Non-Motor", "Truck"]},
                    "EffectiveDate": {
                        "type": "string", 
                        "enum": ["None", "WorkDays"]},
                    "EffectiveTime": {
                        "oneOf": [
                            {
                                "type": "string",
                                "enum": ["None"]},
                            {
                                "type": "string",
                                "pattern": "^([01]?[0-9]|2[0-3]):[0-5][0-9]$"}]},
                    "LowSpeedLimit": {
                        "oneOf": [
                            {
                                "type": "string",
                                "enum": ["None"]},
                            {
                                "type": "string",
                                "pattern": "^[0-9]+$"}]},
                    "HighSpeedLimit": {
                        "oneOf": [
                            {
                                "type": "string",
                                "enum": ["None"]},
                            {
                                "type": "string",
                                "pattern": "^[0-9]+$"}]}},
                "required":["LaneType","RuleIndex","LaneDirection","EffectiveTime","AllowedTransport","EffectiveDate","LowSpeedLimit","HighSpeedLimit"],
                "additionalProperties": false},
            "centerline": {
                "type": "array",
                "items": {
                    "type": "number"}},
            "semantic_polygon": {
                "type": "array",
                "minItems": 3,  
                "items": {
                    "type": "array",
                    "minItems": 3,  
                    "maxItems": 3,
                    "items": {
                        "type": "number"  }}}},
        "required": ["attr_info","centerline","semantic_polygon"],
        "additionalProperties": false }
}
\end{lstlisting}
\end{figure*}

\begin{figure*}[h]
\begin{lstlisting}[language=json, caption=Example of label file., label=label_demo]
{
    "0": {
        "attr_info": {
            "LaneType":         "DirectionLane",
            "RuleIndex":        "1",
            "LaneDirection":    ["GoStraight","TurnLeft"],
            "EffectiveTime":    "None",
            "AllowedTransport": "None",
            "EffectiveDate":    "None",
            "LowSpeedLimit":    "None",
            "HighSpeedLimit":   "None"
        },
        "centerline": [17],
        "semantic_polygon": [
            [6250.473053530053, -23003.147903473426, -51.91421646422327],
            [6250.387053162556, -23003.22814210385,  -53.56106227565867],
            [6249.308139461227, -23004.234772194584, -53.48654436563898],
            [6249.381109470012, -23004.166690932405, -51.82106907669865]
        ]
    },
    "1": {
        "attr_info": {
            "LaneType":         "DirectionLane",
            "RuleIndex":        "2",
            "LaneDirection":    ["GoStraight"],
            "EffectiveTime":    "None",
            "AllowedTransport": "None",
            "EffectiveDate":    "None",
            "LowSpeedLimit":    "None",
            "HighSpeedLimit":   "None"
        },
        "centerline": [16],
        "semantic_polygon": [
            [6249.081411219644,  -23004.446310402054, -53.45673720163109 ],
            [6249.21171480676,   -23004.324736719598, -51.76890653968486 ],
            [6248.1406193206585, -23005.324072389387, -51.694388629665156],
            [6248.0546189531615, -23005.404311019807, -53.37476750060943 ]
        ]
    }
}
\end{lstlisting}
\end{figure*}

\begin{figure*}[h]
    % \ContinuedFloat
    \includegraphics[width = \textwidth]{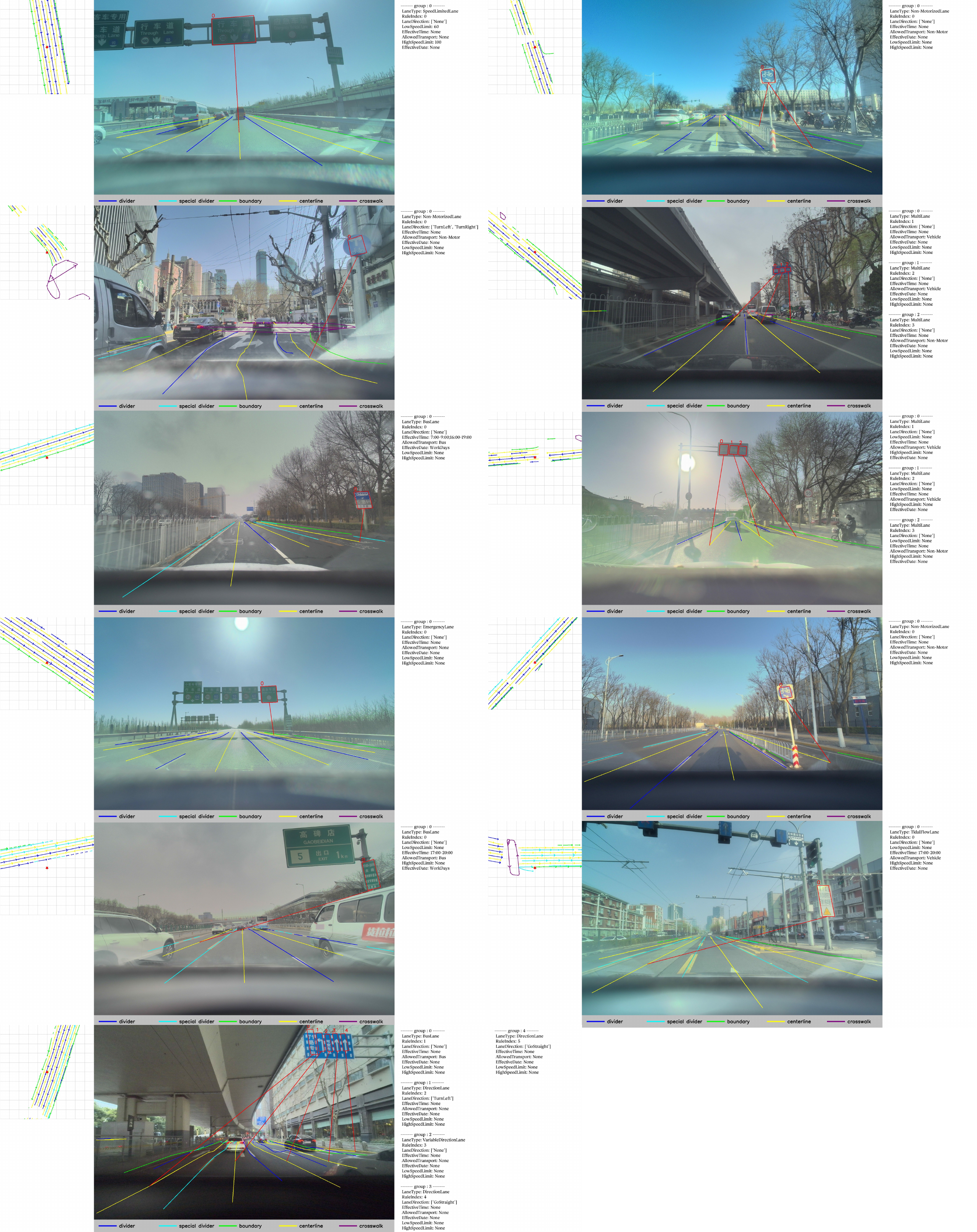}
    \caption{Visualization of MapDR.}
\end{figure*}

\begin{figure*}[h]
    % \ContinuedFloat
    \centering
    \includegraphics[width = 0.95\textwidth]{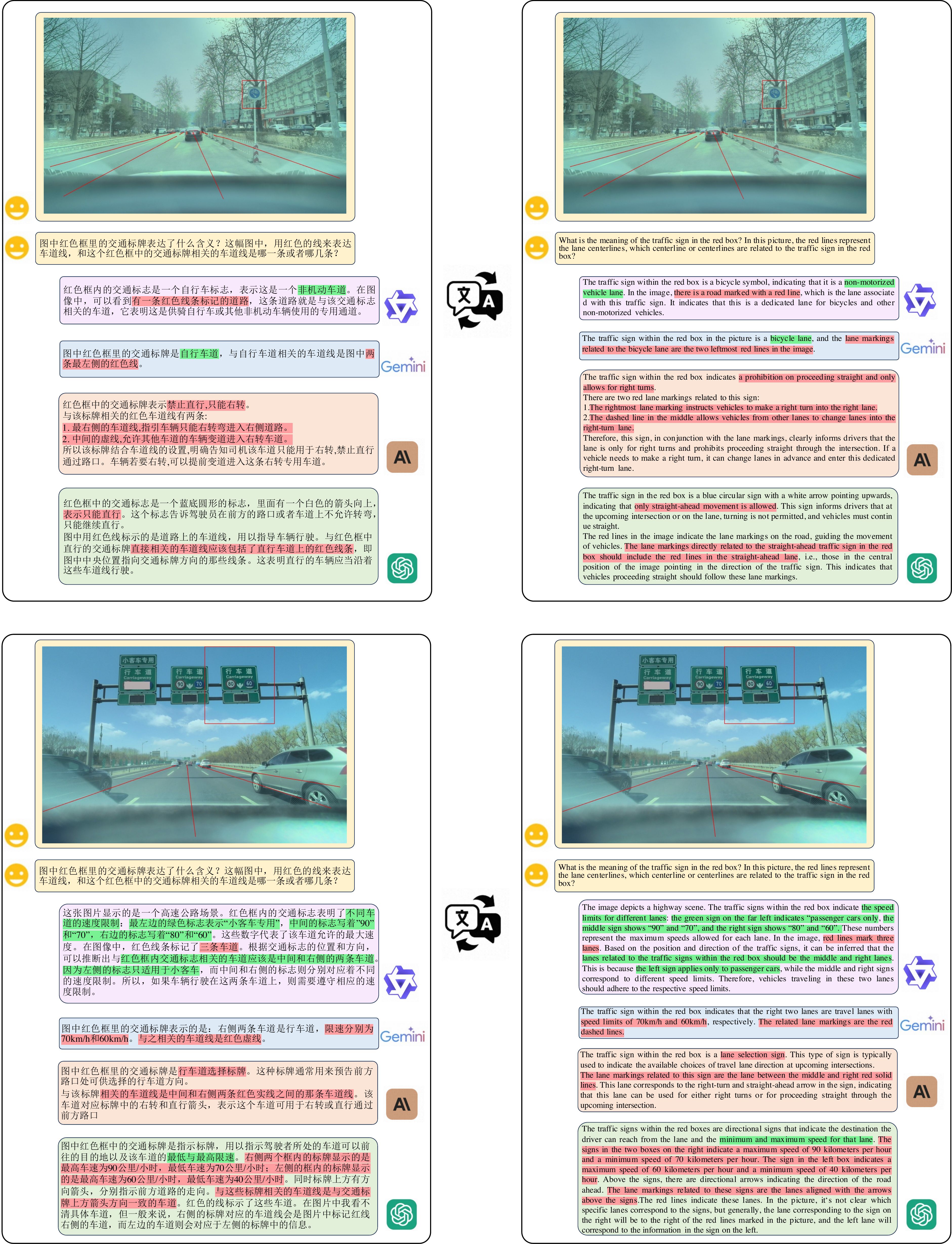}
    \caption{Prompts and answers of MLLMs.}
\end{figure*}

\FloatBarrier

% \clearpage
% \newpage
% {
%     \small
%     \bibliographystyle{ieeenat_fullname}
%     \bibliography{main}
% }

% \end{document}

\end{document}